\pdfoutput=1 
\documentclass[10pt, conference]{IEEEtran} 
\IEEEoverridecommandlockouts
\usepackage{amsmath,amsthm,amssymb,centernot,xspace,microtype}

\newtheorem{definition}{Definition}[section]

\newcommand{\supp}{\text{supp}\xspace}
 \usepackage{algorithm}
 
 \usepackage{algpseudocode}

\usepackage{graphicx,xspace,url}
\graphicspath{{./}{figures/}{comparisons/}}
\usepackage{paralist}

\usepackage{float}
\usepackage{afterpage}

\usepackage[final,inline,nomargin,index]{fixme}
\fxsetup{theme=colorsig,mode=multiuser,inlineface=\itshape,envface=\itshape}
\FXRegisterAuthor{sv}{asv}{Suresh}
\FXRegisterAuthor{sf}{asf}{Sorelle} 
\FXRegisterAuthor{cs}{acs}{Carlos} 
\FXRegisterAuthor{pa}{apa}{Phil}

\newcommand{\acc}{\texttt{acc}}
\newcommand{\ber}{\text{\textsc{ber}}\xspace}

\newcommand{\eobscured}[2]{#1 \setminus_\epsilon #2}

\newcommand{\Obscure}{O}  
\newcommand{\obsval}{x}
\newcommand{\Attr}{W}  
\newcommand{\attrval}{w}
\newcommand{\II}{\text{II}}

\newcommand{\mypara}[1]{\noindent\textbf{#1}}

\begin{document}
\title{Auditing Black-box Models for Indirect Influence\thanks{This research was funded in part by the NSF under grants IIS-1251049, CNS-1302688, IIS-1513651, DMR-1307801, IIS-1633724, and IIS-1633387.}
}

\author{
\IEEEauthorblockN{Philip Adler\IEEEauthorrefmark{1}, Casey Falk\IEEEauthorrefmark{1}, Sorelle A. Friedler\IEEEauthorrefmark{1}, Gabriel Rybeck\IEEEauthorrefmark{1},\\ Carlos Scheidegger\IEEEauthorrefmark{2}, Brandon Smith\IEEEauthorrefmark{1}, and Suresh Venkatasubramanian\IEEEauthorrefmark{3}}
\IEEEauthorblockA{\IEEEauthorrefmark{1}
Dept. of Computer Science, Haverford College, Haverford, PA, USA\\
\url{padler1@haverford.edu}, \url{caseyfalk94@gmail.com}, \url{sorelle@cs.haverford.edu}, \url{grybeck@gmail.com}, \url{bsmith8108@gmail.com}
}
\IEEEauthorblockA{\IEEEauthorrefmark{2}
Dept. of Computer Science, University of Arizona, Tucson, AZ, USA\\
\url{cscheid@cscheid.net}
}
\IEEEauthorblockA{\IEEEauthorrefmark{3}
Dept. of Computer Science, University of Utah, Salt Lake City, UT, USA\\
\url{suresh@cs.utah.edu}
}
}

\maketitle

\begin{abstract}

  Data-trained predictive models see widespread use, but for the most part they are used as \emph{black boxes} which output a prediction or score. It is therefore hard to acquire a deeper understanding of model behavior, and in particular how different features influence the model prediction. This is important when interpreting the behavior of complex models, or asserting that certain problematic attributes (like race or gender) are \emph{not} unduly influencing decisions.

  In this paper, we present a technique for \emph{auditing} black-box models, which lets us study the extent to which existing models take advantage of particular features in the dataset, without knowing how the models work. Our work focuses on the problem of \emph{indirect influence}: how some features might indirectly influence outcomes via other, related features. As a result, we can find attribute influences even in cases where, upon further direct examination of the model, \emph{the attribute is not referred to by the model at all.}

Our approach does not require the black-box model to be retrained. This is important if (for example) the model is only accessible via an API, and contrasts our work with other methods that investigate feature influence like feature selection.
 We present experimental evidence for the effectiveness of our procedure using a variety of publicly available datasets and models. We also validate our procedure using techniques from interpretable learning and feature selection, as well as against other black-box auditing procedures.
\end{abstract}


\section{Introduction}

Machine-learning models now determine and control an increasing number of real-world decisions, from sentencing guidelines and parole hearings~\cite{propublica} to predicting the outcome of chemical experiments~\cite{drpNature}. These models, powerful as they are, tend to also be opaque. 
%
%
This presents a challenge. How can we \emph{audit} such models to understand why they make certain decisions? 
As conscientious model creators, we should want to know the extent to which a specific feature contributes to the accuracy of a model. 
As outside auditors, trying to understand a system can give us an understanding of the model's priorities and how it is influenced by certain features. This may even have legal ramifications: by law for example, decisions about hiring cannot be influenced by factors like race, gender or age. 

As model creators, we could build interpretable models, either by explicitly using interpretable structures like decision trees, or by building shadow models that match model outputs in an interpretable way. In this work, we are interested in auditing a \emph{black box} model from the outside (because the model is proprietary, accessible only through an API, or cannot be modified).




\subsection{Direct and Indirect Influence}
\label{sec:proxies}

Much of the modern literature on black-box auditing (See Section~\ref{sec:altern-appr} for details) focuses on what we call \emph{direct} influence: how does a feature (or a group of features) directly affect the outcome? This is quantified by replacing the feature (or group) by random noise and testing how model accuracy deteriorates. In this paper, we focus on the different and more subtle issue of \emph{indirect} influence. 


Consider trying to verify that racial considerations did \emph{not} affect an automated decision to grant a housing loan. We could use a standard auditing procedure that declares that the \texttt{race} attribute does not have an undue influence over the results returned by the algorithm. Yet this may be insufficient. In the classic case of \emph{redlining} \cite{redlining}, the decision-making process explicitly excluded race, but it used \texttt{zipcode}, which in a segregated environment is strongly linked to race. Here, race had an \emph{indirect} influence on the outcome via the zipcode, which acts as a \emph{proxy}. 

Note that in this setting, race would not be seen as having a direct influence (because removing it doesn't remove the signal that it provides). Removing both race and zipcode jointly (as some methods propose to do) reveals their combined influence, but also eliminates other task-specific value that the zipcode might signal independent of race, as well as leaving unanswered the problem of \emph{other} features that race might exert indirect (and partial) influence through.




\subsection{Our Work}
\label{sec:our-work}

In this paper, we study the problem of auditing black box models for indirect influence. In order to do this, we must find a way to capture information flow from one feature to another. We take a learning-theoretic perspective on this problem, which can be summarized via the principle, first enunciated in \cite{2015_kdd_disparate_impact} in the context of certifying and removing bias in classifiers:
\begin{quote}
  \emph{the information content of a feature can be estimated by trying to
    \emph{predict} it from the remaining features}.
\end{quote}

How does this allow us to correctly quantify the influence of a feature in the presence of proxies? Let us minimally modify the data so that the feature can no longer be predicted from the remaining data. The above principle then argues that we have fully eliminated the influence of this feature, both directly and in any proxy variables that might exist. If we now test our model with this \emph{obscured} data set, any resulting drop in prediction accuracy can be attributed directly to information from the eliminated feature. 

Our main contributions in this work include
\begin{itemize}
\item A technique to \emph{obscure} (fully and partially) the influence of a feature on an outcome, and a theoretical justification of this approach.
\item A method for quantifying indirect influence based on a differential analysis of feature influence before and after obscuring. 
\item An experimental validation of our approach on a number of public data sets, using a variety of models. 
\end{itemize}

\section{Conceptual Context}
\label{sec:altern-appr}

Work on black-box auditing intersects with a number of related areas in machine learning (and computer science at large), including directions coming from privacy preservation, security, interpretability, and feature selection. We tease out these connections in more detail in Section~\ref{sec:related-work}.

Here we outline how we see our work in the specific context of the prior literature on auditing black box machine learning models. Modern developments in this area can be traced back to Breiman's work on random forests \cite{breiman2001random}, and we highlight two specific recent related works. 
Henelius et al. \cite{Henelius2014BlackBox} propose looking at variable sets and their influence by studying the consistency of a black-box predictor when compared to a random permutation of a set. Datta et al. \cite{datta2016} provide a generalization of this idea, linking it to game-theoretic notions of influence and showing that different choices of probability spaces and random variables yield a number of different interesting auditing measures. These two papers fundamentally hinge on the notion of associating each input value with an \emph{intervention distribution}. These intervention distributions can be easily shown (in distribution) to obscure attributes. Our work, on the other hand, will audit black boxes by providing, for any given input point, an intervention that is \emph{deterministic}, while guaranteeing (in some settings, see Section \ref{sec:anova}) that the attributes are still obscured over the entire distribution. Our intervention preserves more of the signal in the dataset, and -- crucially in some settings -- naturally preserves indirect influences of proxy variables. As we will show in the experiments (see Section \ref{sec:comparison}), the technique of Henelius et al. cannot detect proxy variables, and although Datta et al. can use some of their measures to detect proxy variables, their attribute rankings generally do not reflect the proxy relationships.

Our methods draw heavily on ideas from the area of \emph{algorithmic fairness}. The process by which we eliminate the influence of a feature uses ideas from earlier work on testing for disparate impact \cite{2015_kdd_disparate_impact}. Again, a key difference is that we no longer have the ability to retrain the model, and we \emph{quantify} the influence of a feature rather than merely eliminating its influence.


\section{Indirect Influence}
\label{sec:problem}
Let $f \colon \mathbb{X} \to \mathbb{Y}$ be a black-box classification function, where $\mathbb{X} \subset \mathbb{X}^{(1)} \times \mathbb{X}^{(2)} \ldots \mathbb{X}^{(d)}$ is a $d$-dimensional feature space with the $i^{th}$ coordinate drawn from the domain  $\mathbb{X}^{(i)}$  and $\mathbb{Y}$ is the domain of outcomes. For example, $\mathbb{X} = \mathbb{R}^d$ and $\mathbb{Y} = \{-1,1\}$ for binary classification on Euclidean vectors. Fix a data set $(X, Y) = \{(X_i, y_i)\}\subset \mathbb{X} \times \mathbb{Y}, 1 \le i \le n$ and let $X_i = (x_{i1}, x_{i2}, \ldots, x_{id})$, where $x_{ij} \in \mathbb{X}_j$ for all $i$.  We denote the accuracy of prediction as $\acc(X, Y, f)$. For example, $\acc(X, Y, f) = \frac{1}{n}\sum \mathbf{1}_{y_i \ne f(X_i)}$ is the standard misclassification error.

We wish to quantify the indirect influence of a feature $j$ on the outcome of classification. The typical approach to doing this in the auditing literature is to perturb the $j^{th}$  feature $x_{ij}$ of each $X_i$ in some way (usually by random perturbation), obtaining a modified data set $X_{-j}$. Then the influence of $j$ can be quantified by measuring the difference between $\acc(X, Y, f)$ and $\acc(X_{-j}, Y, f)$ (note that $f$ is not retrained on $X_{-j}$). 

Unfortunately, randomly perturbing features can disrupt indirect influence in a number of ways. Firstly, random perturbations could also remove useful task-related information in proxy features that would degrade the quality of classification. Secondly, this prevents us from cleanly quantifying the \emph{relative} effect of the feature being perturbed on related proxy variables.

We propose a different approach. We will still perturb a data set $X$ to eliminate the direct and indirect influence of feature $j$, and measure the change in accuracy of prediction as before. However, we will do this perturbation in a directed and deterministic manner, organized around the question: ``\emph{can we predict the value of feature $j$ from the remaining features}?'' Intuitively, if we cannot, then we know that we have correctly eliminated the influence of $j$. Moreover, if we do this perturbation ``minimally,'' then we have changed the data as little as possible in the process. We will say in this case that we have \emph{removed} feature $j$ from the data set and have  \emph{obscured} its influence on $X$. 

\subsection{Obscuring data with respect to a feature}
\label{sec:obscuring-features}
We start by defining the error measure we will use to test predictability. Rather than the standard misclassification rate, we will use the well-known \emph{balanced error rate} measure that is more sensitive to class imbalance. This is important if a feature has significant skew in the values it takes. Let $\supp(Y) = \{y \in \mathbb{Y}|y \in Y\}$ be the set of elements of $\mathbb{Y}$ that appear in the data. 

\begin{definition}[BER]
Let $f : \mathbb{X} \to \mathbb{Y}$ be a classifier, and let $(X,Y) = \{(X_i, y_i)\}$ be a set of examples. 
The \emph{balanced error rate} \ber of $f$ on $(X,Y)$ is the (unweighted) average class-conditioned error of $f$:
\[ \ber(X, Y, f) = \frac{1}{|\supp(Y)|}\left( \sum_{j \in \supp(Y)}\frac{\sum_{y_i = j} \mathbf{1}_{f(X_i) \ne j}}{|\{i \mid y_i = j\}|} \right)\]
\end{definition}

A feature $i$ has been removed from a data set if we can no longer predict that feature from the remaining data. This motivates the following definition. Let $X^{(i)} = (x_{1i}, x_{2i}, \ldots, x_{ni})$ denote the column corresponding to the $i^{th}$ feature. 

\begin{definition}[$\epsilon$-obscure]
\label{def:obscure}
 We define $\eobscured{X}{\mathbb{X}_i}$ as the \emph{$\epsilon$-obscure} version of $X$ with respect to feature $\mathbb{X}_i$ if $X^{(i)}$ cannot be predicted from $\eobscured{X}{X_i}$.  I.e., if, for all functions $f \colon \mathbb{X}\setminus \mathbb{X}_i \to \mathbb{X}_i$,
\[ \ber(\eobscured{X}{\mathbb{X}_i}, X^{(i)}, f) > \epsilon \]
\end{definition}

We can now define a measure of influence for a feature. 

\begin{definition}[(indirect) influence]
The \emph{indirect influence} \II(i) of a feature $i$ on a classifier $f$ applied to data $(X, Y)$ is the difference in accuracy when $f$ is run on $X$ versus when it is run on $\eobscured{X}{\mathbb{X}_i}$:
\[ \II(i) = \acc(X, Y, f) - \acc(\eobscured{X}{\mathbb{X}_i}, Y, f) \]
\end{definition}


\mypara{Notes}
The definition of obscurity we use here is adapted from \cite{2015_kdd_disparate_impact}, but applied to any feature, rather than just ``protected'' ones. In what follows, we will typically treat $\epsilon$ as large (say above 0.5 for binary classification). 


\section{Computing Influence}
\label{sec:gradient_auditing}

In this section, we will introduce a method we call \emph{gradient feature auditing} (GFA) to estimate indirect influence. Using it, we compute the influence of each feature of the data and order the features based on their influence. This GFA algorithm works feature-by-feature: in order to compute $\eobscured{X}{\mathbb{X}_i}$, we apply an \emph{obscuring procedure} to each feature $j \ne i$ in $\mathbb{X}$. Because we operate one feature at a time, we cannot guarantee that all influence can be removed (and therefore estimated). However, we will show that the feature-level procedure is theoretically sound and in fact generalizes the standard ANOVA test for null hypothesis testing. 

Let us start with a simple case: when the feature $\Attr = \mathbb{X}_j$ to be obscured is numerical, and the feature $\Obscure = \mathbb{X}_i$ we are removing is categorical. Let $\Attr_\obsval = \Pr(\Attr \mid \Obscure = \obsval)$ denote the marginal distribution on $\Attr$ conditioned on $\Obscure = \obsval$ and let the cumulative distribution be $F_\obsval(\attrval) = \Pr(\Attr \ge \attrval \mid \Obscure = \obsval)$.

Define the \emph{median} distribution $A$ such that its cumulative distribution $F_A$ is given by 
$F_A^{-1}(u) = \text{median}_{\obsval \in \Obscure} F_\obsval^{-1}(u) \mbox{ .}$
In \cite{2015_kdd_disparate_impact} it was shown that if we modify the distribution of $W$ to match $A$ by ``moving'' values of $W$ so as to mimic the distribution given by $A$, then $\Obscure$ is maximally obscured, but $\Attr$ also minimally changes, in that $A$ also minimizes the function $\sum_{\obsval \in \Obscure} d(\Attr_\obsval, A)$ where $d(\cdot, \cdot)$
was the earthmover distance \cite{emd} between the distributions using $\ell_2$ as the base
metric. We call this procedure \texttt{ObscureNumerical}. 

This procedure does not work if features to be obscured and removed are not numerical and categorical respectively.
We now describe procedures to address this issue. 

\subsection{Removing numerical features}
In order to remove a numerical feature, we must first determine what aspects of the number itself should be removed.  In an optimal setting, we might remove the entirety of the number by considering its binary expansion and ensuring that no bit was recoverable.  However, when working with most numerical data, we can safely assume that only the higher order bits of a number should be removed.  For example, when considering measurements of scientific phenomena, the lower order bits are often measurement error.

Thus, we bin the numerical feature and use the bins as categorical labels in the previously described obscuring procedure.  Bins are chosen using the Freedman-Diaconis rule for choosing histogram bin sizes \cite{freedman1981histogram}.

\subsection{Obscuring categorical features}

Our procedure relies on being able to compute cumulative density functions for the feature $\Attr$ being obscured.
If it is categorical, we no longer have an ordered domain on which to
define the cumulative distributions $F_\attrval$. However, we do have a base metric:
the exact metric $\boldsymbol{1}$ where $\boldsymbol{1}(\obsval, \attrval) = 1 \iff \obsval = \attrval$. 
We can therefore define $A$ as before, as the distribution minimizing the
function $\sum_{\obsval \in \Obscure} d(\Attr_\obsval, A)$.  We observe that the earthmover distance between any two distributions over the exact metric has a particularly simple form. Let $p(\attrval), q(\attrval), \attrval \in \Attr, \sum p(\attrval) = \sum q(\attrval) = 1$ be two distributions over  $\Attr$. Then the earthmover distance between $p$ and $q$ with respect to the exact metric $\boldsymbol{1}$ is given by $d(p, q) = \|p - q\|_1$
Therefore the desired minimizer $A$ can be found by taking a component-wise median for each value $\attrval$. In other words, $A$ is the distribution such that $p_A(\attrval) = \text{median}_\attrval \Attr_\obsval(\attrval)$.  Once such an $A$ has been computed, we can find the exact repair by computing the earthmover distance (via min-cost flows)\footnote{This is a straightforward application of the standard min-cost flow problem; we defer a detailed description to the full version of this paper.} between each $\Attr_\obsval$ and $A$. This results in fewer changes than merely changing values arbitrarily.

We must create the obscured version of $\Attr$, denoted $\hat{\Attr}$.  Let $\hat{\Attr}_{\attrval, \obsval}$ be the partition of $\hat{\Attr}$ where the value of the obscured feature is $\attrval$ and the value of the removed feature is $\obsval$.  We must create $\hat{\Attr}$ so as to ensure that $|\{\hat{\Attr} | \Obscure = \obsval \}| \in \mathbb{Z}$ for all values of $\obsval \in \Obscure$.  We set
$ |\hat{\Attr}_{\attrval, \obsval}| = \lfloor p_A(\attrval) \cdot |\{\Attr | \Obscure = \obsval\}| \rfloor$.
Letting $d(\attrval) = |\hat{\Attr}_{\attrval, \obsval}|$ for all $\attrval \in \Attr$ gives the node demands for the circulation problem between $\Attr_\obsval$ and $A$, where supplies are set to the original counts $d(\attrval) = - |\Attr_{\attrval, \obsval}|$.  Since $|\hat{\Attr}_{\attrval, \obsval}| \leq |\Attr_{\attrval, \obsval}|$, an additional \emph{lost observations node} with demand $|\Attr_{\attrval, \obsval}| - |\hat{\Attr}_{\attrval, \obsval}|$ is also added.  The flow solution describes how to distribute observations at a per category, per obscured feature value level.  Individual observations within these per category, per feature buckets can be distributed arbitrarily.  The observations that flow to the lost observations node are  distributed randomly according to distribution $A$. We call this procedure \texttt{ObscureCategorical}.

Using the categorical or numerical obscuring procedure appropriately depending on the data yields our procedure for computing $\eobscured{X}{\mathbb{X}_i}$.

%

\noindent\textbf{Notes.} This description assumes that we want to remove \emph{all} effects of a variable in order to measure its influence. However, how the influence changes as we remove its effect is also interesting. Indeed, this is why we we refer to the overall process as a \emph{gradient} feature audit.  To that end, all of the algorithms above can be adapted to allow for a \emph{partial} removal of a variable. On a scale of $0$ to $1$ where $0$ represents the original data, and $1$ represents a full removal, we can remove a \emph{fractional} part of the influence by allowing the individual conditional distributions to move  \emph{partly} towards each other.

While the process above produces a score, the induced ranking is also useful, especially when we compare our results to those produced by other auditing methods, where the score itself might not be directly meaningful. We illustrate this further in Section~\ref{sec:experiments}.

\subsection{Obscuring, ANOVA, and the $F$-test}
\label{sec:anova}

We now provide a theoretical justification for our obscuring procedure. Specifically, we
show that if the feature $\Attr$ being obscured has Gaussian conditionals, then
our obscuring procedure will create a data set on which the
F-test \cite{StatsBook} will fail, which means that the null hypothesis (that the
conditionals are now identical) will not be invalidated. Thus, our procedure can
be viewed as a generalization of ANOVA.

Consider a data set $D$ consisting of samples $(x, y)$, where $y$ is a class label
that takes the values $-1,1$ and the $x$ are drawn from univariate Gaussians with
different means and a shared variance. Specifically, 
$ \Pr(x | y = i) = \mathcal{N}(v_i, \sigma^2) $.
We assume the classes are balanced: $\Pr(y = -1) = \Pr(y = 1)$.

Let $\tilde{x}$ be the obscured version of $x$. It is easy to show
that\footnote{This follows from the fact that the earthmover distance between
  two distributions on the line is the $\ell_1$ difference between their
  cumulative density functions. In this case it means that the earthmover
  distance is precisely the distance between the means.}, in this case,
the obscuring procedure will produce values following
the distribution $p(\tilde{x} | y = i) = \mathcal{N}(1/2 (v_{-1} + v_1), \sigma^2)$.

We apply the $F$-test to see if we can tell apart the two conditional
distributions for the two different values of $y$. Let $S_y = \{ x \mid
(x,y) \in D\}$. The test statistic $F$ is the
ratio of the between-group sample variance and the in-group sample
variance. Set $\mu = (v_{-1} + v_1)/2$ and $\delta = (\mu - v_{-1})^2 = (\mu -
v_1)^2$. Thus
\[ F = \frac{n \delta}{(1/2) \sum_{x \in S_{-1}}(x - v_{-1})^2 + \sum_{x \in
    S_1}(x - v_1)^2} \]
We note that $\sum_{x \in S_1}(x - v_1)^2$ has expectation $n/2 \sigma^2$, (and
so does the corresponding expression related to $S_{-1}$). Using the plug-in
principle, we arrive at an estimator for the $F$ statistic:
$ F = \frac{\delta}{\sigma^2}$
This is the traditional expression for a two-variable, one-way ANOVA: $\delta$
is a measure of the variance that is explained by the group, and $\sigma^2$ is
the unexplained variance. 

We apply this to the obscured distribution. From the remarks above, we
know that the conditional distributions for $y = 0,1$ are identical Gaussians
$N(\mu, \sigma)$. We now show that the positive parameter $\delta$ is
concentrated near zero as the number of samples increases. 

Let $x_1, \ldots, x_n$ be samples drawn from the conditional distribution for $y
= 0$, and similarly let $y_1, \ldots, y_n$ be drawn from the distribution
conditioned on $y = 1$. Set $X = \frac{1}{n} \sum x_i$ and $Y = \frac{1}{n}\sum
y_i$. Note that $E[X] = E[Y] = \mu$, and so $E[X-Y] = 0$. 

Let $\hat{\delta} = (X - Y)^2$. We first observe that 
\begin{align*}
   |X-Y| &\le |X - E[X]| + |E[X] - E[Y]| + |Y - E[Y]| \\
  &= 2|X - E[X]|
\end{align*}
because $X$ and $Y$ are identically distributed. 
Therefore,
\begin{align*}
  \Pr(\hat{\delta} \ge \epsilon^2) &\le 4\Pr(|X - E[X]|^2 \ge \epsilon^2)\\
                                   &\le 4 \exp(-n\epsilon^2/2\sigma^2)
\end{align*}
by standard Hoeffding tail bounds \cite{randomizedalgos}. 
Therefore, with $\log n/\epsilon^2$ samples, the $F$-test statistic is with high
probability less than $\epsilon^2$, and thus the null hypothesis (that the
distributions are identical) will not be invalidated (which is equivalent to
saying that the test cannot distinguish the two conditional distributions).

\section{Experiments}
\label{sec:experiments}
In order to evaluate the introduced gradient feature auditing (GFA) algorithm, we consider experiments on five data sets, chosen to balance easy replicability with demonstration on domains where these techniques are of practical interest.  Datasets and GFA code are available online.\footnote{\url{https://github.com/cfalk/ BlackBoxAuditing}}

\mypara{Synthetic data.}  We generated $6,000$ items with $3,000$  assigned to each of two classes. Items have five features. Three features directly encode the row number $i$: feature A is $i$, B is $2i$, and $C$ is $-i$.  There is also a random feature and a constant feature.  We use a random $\frac{2}{3}:\frac{1}{3}$ training-test split.

\mypara{Adult Income and German Credit data.} We  consider two commonly used data sets from the UC Irvine machine learning repository\footnote{\url{https://archive.ics.uci.edu/ml/datasets.html}}.  The first is the Adult Income data set consisting of 48,842 people, each with 14 descriptive attributes from the US census and a classification of that person as making more or less than \$50,000 per year.
We use the training / test split given in the original data.  The second is the German Credit data set consisting of 1000 people, each with 20 descriptive attributes and a classification as having good or bad credit.  We use a random $\frac{2}{3}:\frac{1}{3}$ training-test split on this data and the two data sets described below.

\mypara{Recidivism data.}  The Recidivism Prediction data set is taken from the National Archive of Criminal Justice Data\footnote{\url{http://doi.org/10.3886/ICPSR03355.v8}} and contains information on 38,624 prisoners sampled from those released from 15 states in 1994 and tracked for three years.  The full data includes attributes describing the entirety of the prisoners' criminal histories before and after release as well as demographic data.  For this work, we processed the data additionally following the description of the processing performed in a previous study on interpretable models \cite{Zeng2015Interpretable}.  This processing resulted in 10 categorical attributes.  The prediction problem considered is that of determining whether a released prisoner will be rearrested within three years.

\mypara{Dark Reactions data.} The Dark Reactions data \cite{drpNature} is a set of $3,955$ historical hydrothermal synthesis experiments aiming to produce inorganic-organic hybrid materials. $273$ attributes indicate aggregates of atomic and ionic properties of the inorganic components and semi-empirical properties of the organic components. The classification variable indicates whether the reaction produced an ionic crystal.


\mypara{Models.}
We built two models that are notoriously opaque to simple examination; SVMs\footnote{Implemented using Weka's version 3.6.13 SMO: \url{http://weka.sourceforge.net/doc.dev/weka/classifiers/functions/SMO.html}} \cite{Hastie1998} and feedforward neural networks (FNNs).\footnote{Implemented using TensorFlow version 0.6.0: \url{https://www.tensorflow.org/}}  We also include C4.5 decision trees\footnote{Implemented using Weka's version 3.6.13 J48: \url{http://weka.sourceforge.net/doc.dev/weka/classifiers/trees/J48.html}} \cite{Quinlan1993C4.5} 
so that the audited results can be examined directly in comparison to the models themselves.

The FNN on the synthetic data was trained using a \texttt{softmax} input-layer for 100 epochs with a batch-size of 500 and a learning-rate of 0.01; no hidden layer was involved.  The Adult Income FNN model was trained using a single \texttt{softmax} input-layer for 1000 epochs using a batch-size of 300 and a learning-rate of 0.01; no hidden layer was involved.  The German Credit data FNN model was trained similarly to the Adult model, except using a batch-size of 700.  The Dark Reaction data FNN model was trained using \texttt{tanh} activations for the input layer and \texttt{softmax} activations in a fully connected hidden layer of 50 nodes; it was trained using a batch-size of 300 for 1500 epochs with a modified learning rate of 0.001.  The Recidivism data set FNN was trained using \texttt{softmax} activations on a fully connected hidden layer of 100 nodes.  The model was trained using a batch size of 500 and 100 epochs.

\mypara{Auditing using test data.}  Our claim in this work is that we can obscure a data set with respect to a feature in order to test its influence. To test this, we could also retrain our classifier on obscured data and compare the resulting outcomes on test data.  We have run this experiment on the synthetic data and found the resulting scores to be very similar, demonstrating that even though our obscuring process applies after training, it is still effective at removing the influence of a feature. We defer detailed experiments to an extended version of this work.

\subsection{Black-box feature auditing}
   
\begin{figure*}[!hp]
\begin{center}
\begin{tabular}{ccc}
\includegraphics[width=2in]{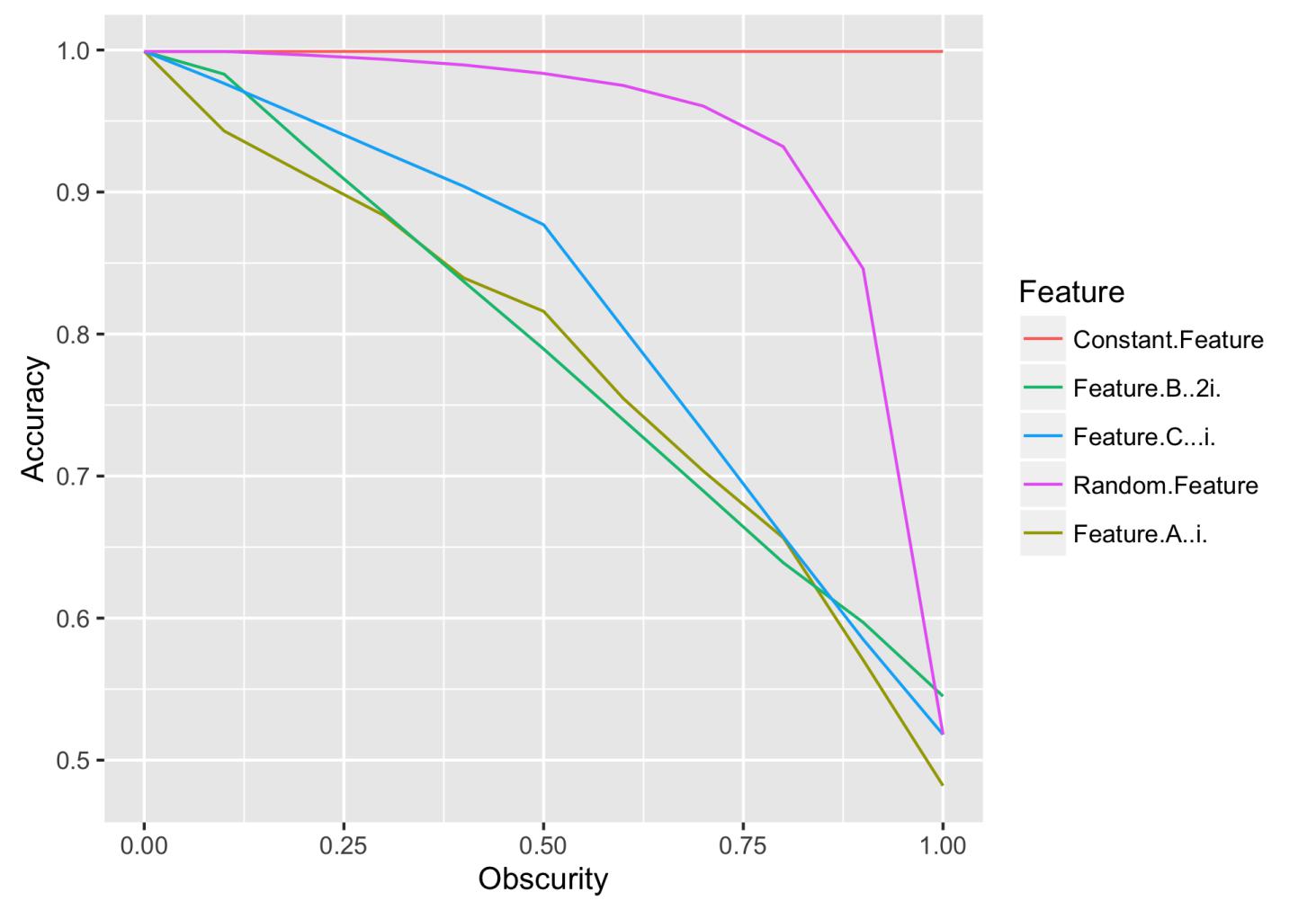} & \includegraphics[width=2in]{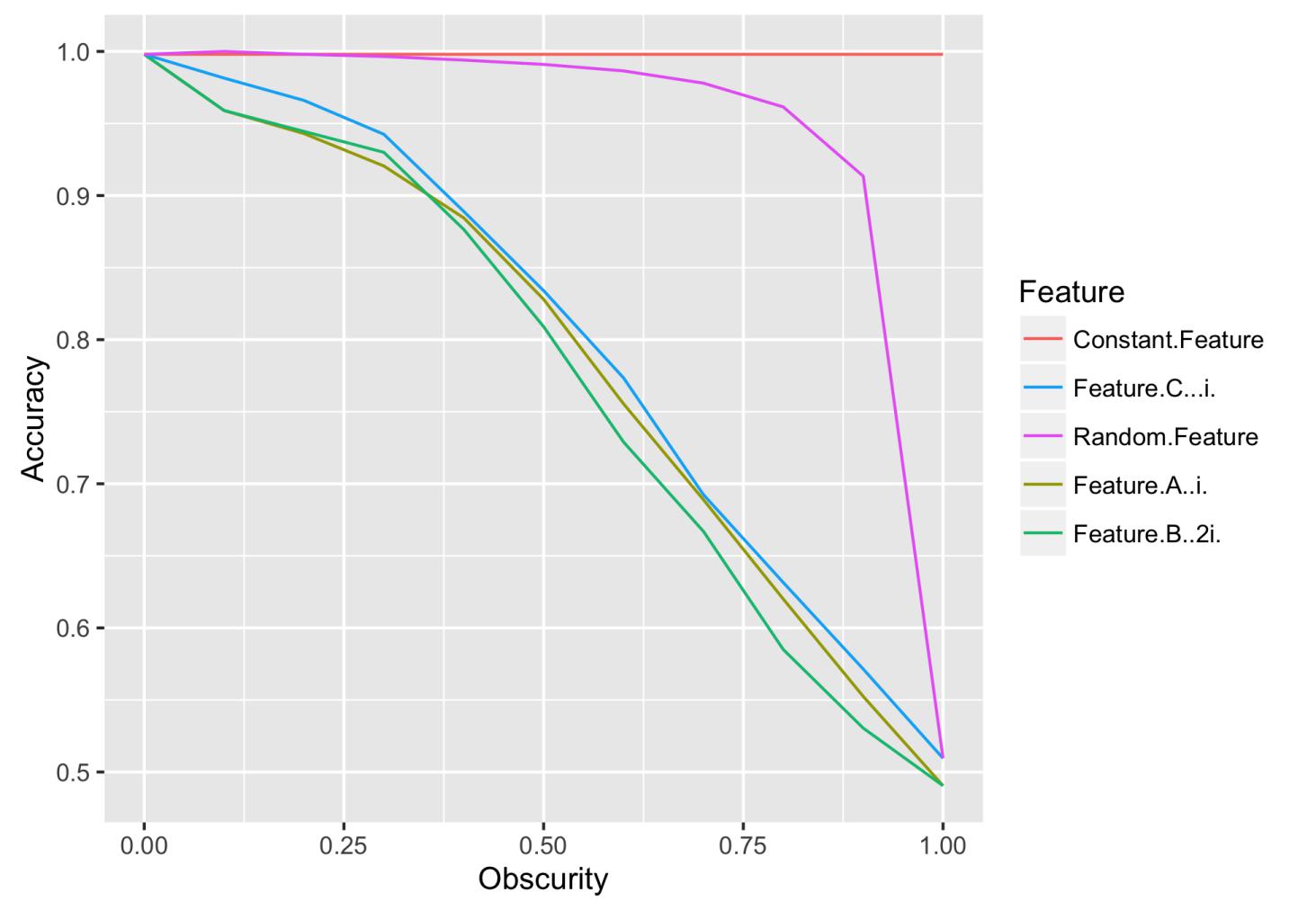} & \includegraphics[width=2in]{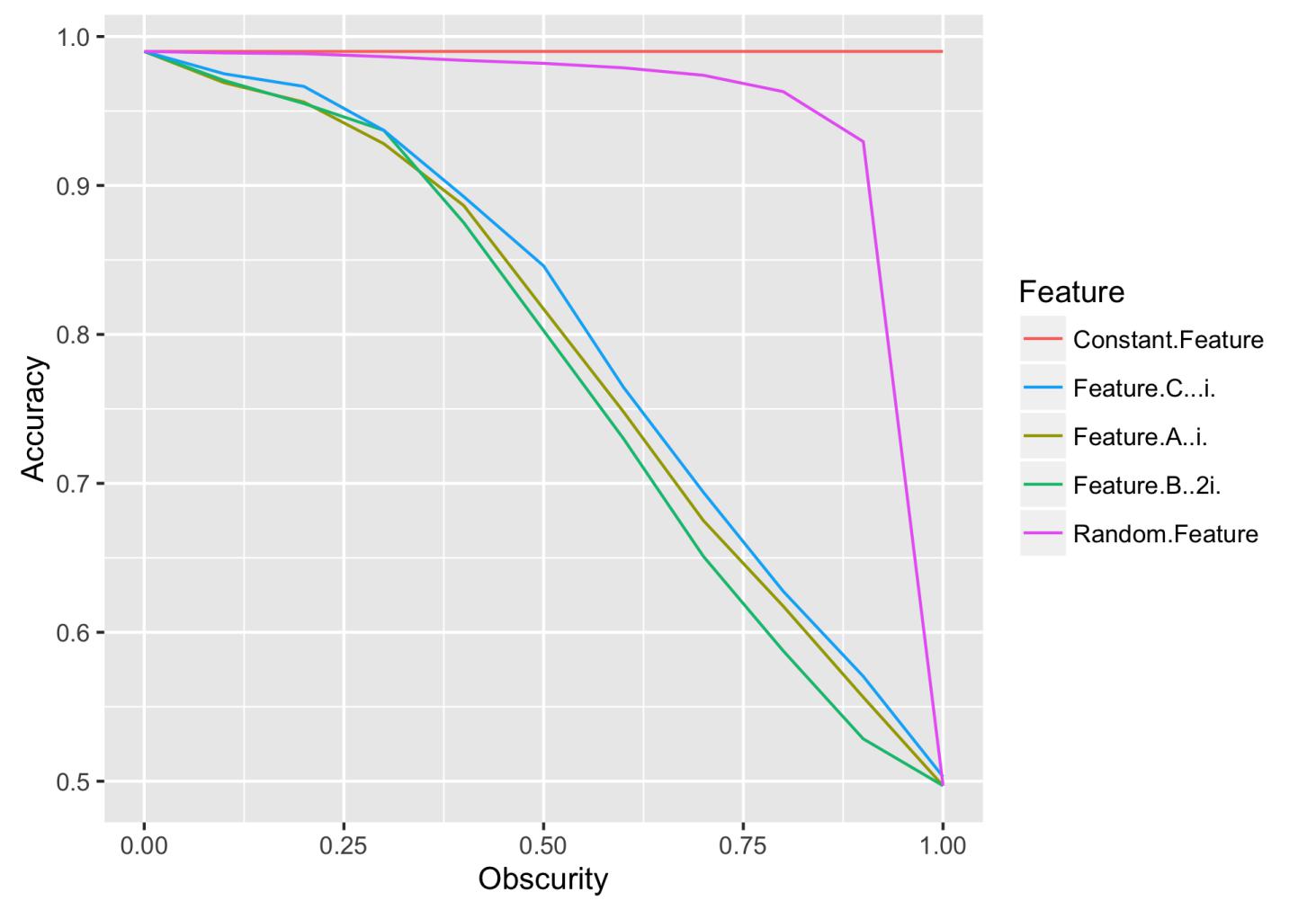}\\
\includegraphics[width=2in]{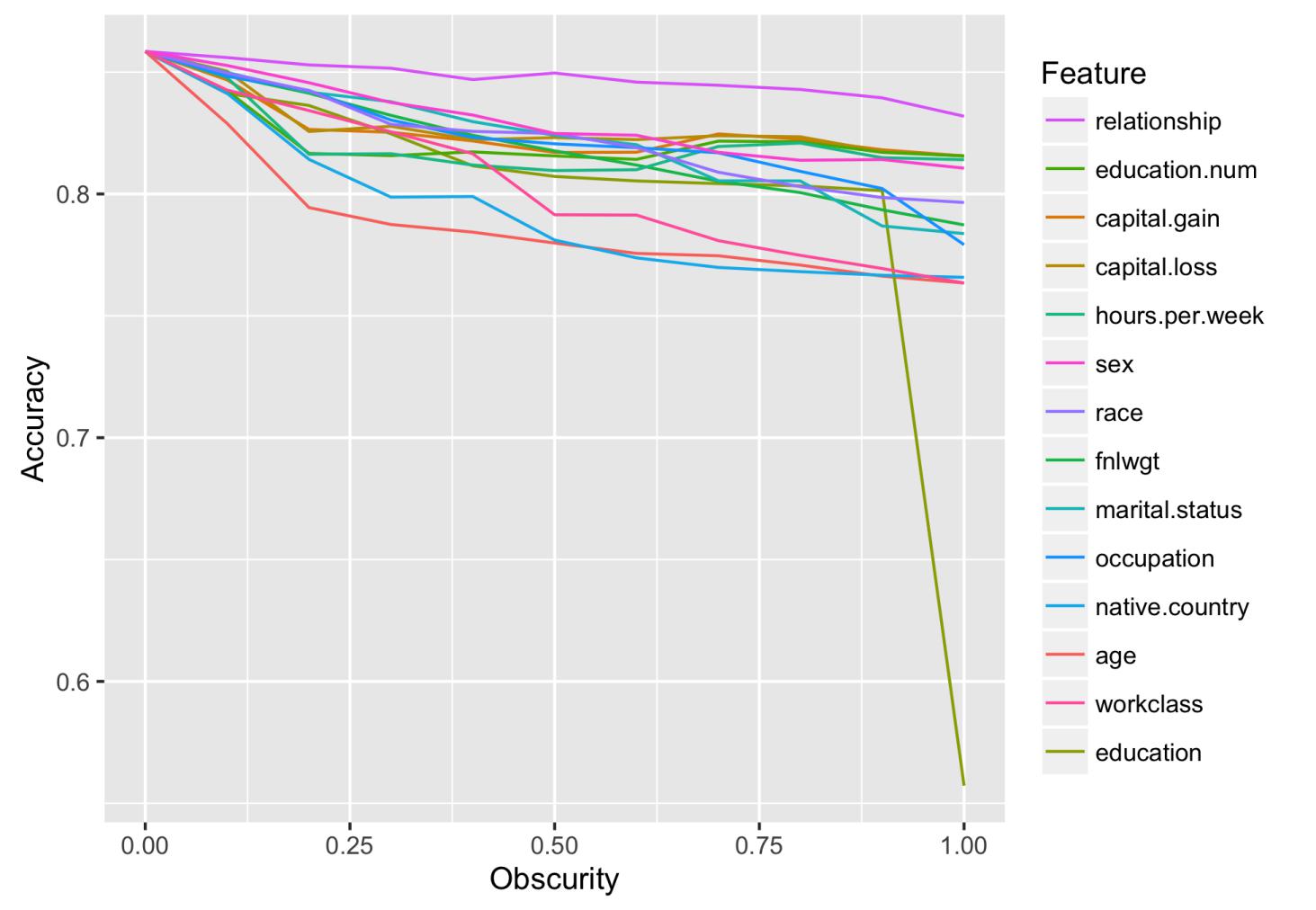} & \includegraphics[width=2in]{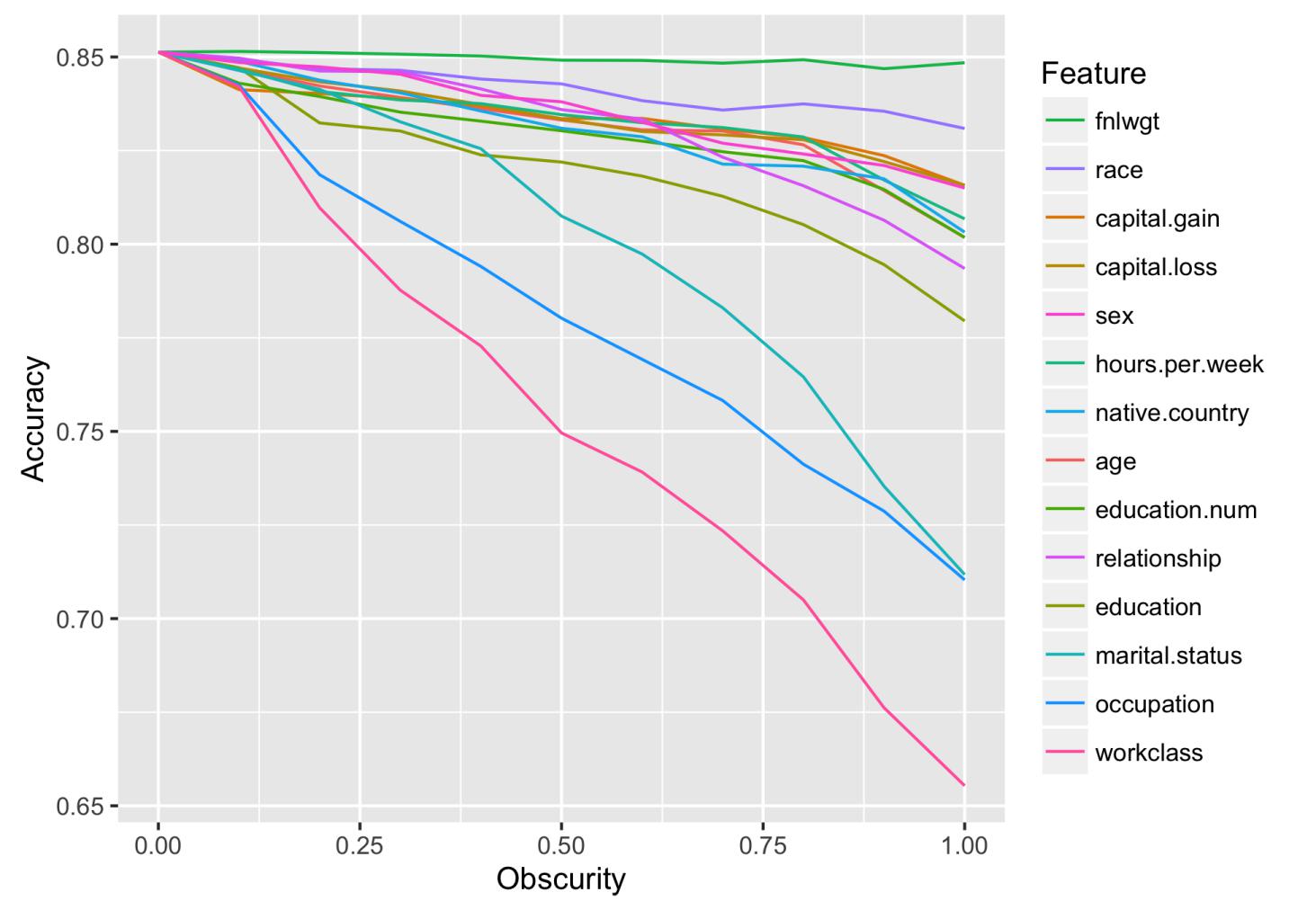} & \includegraphics[width=2in]{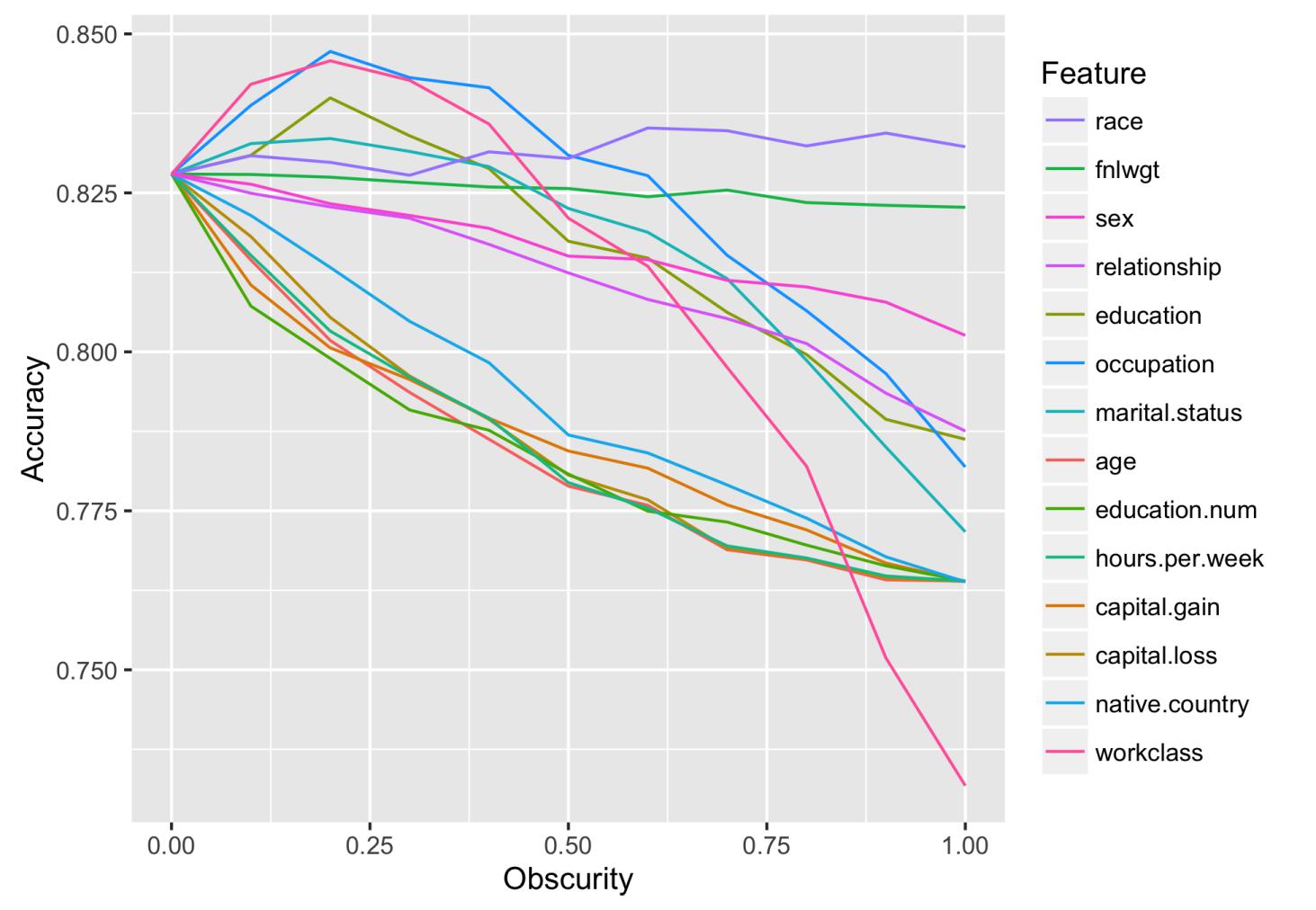}\\
\includegraphics[width=2in]{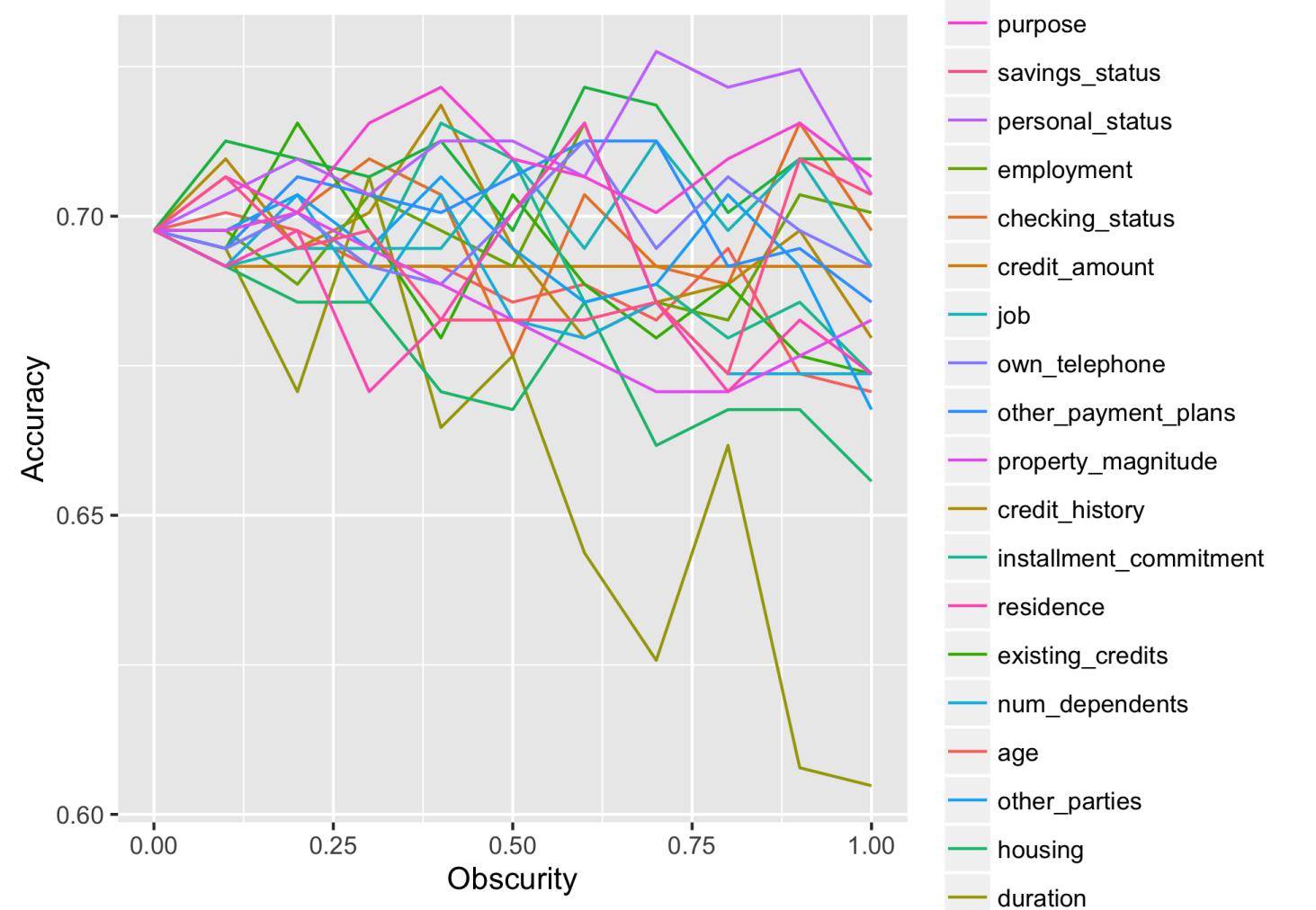} & \includegraphics[width=2in]{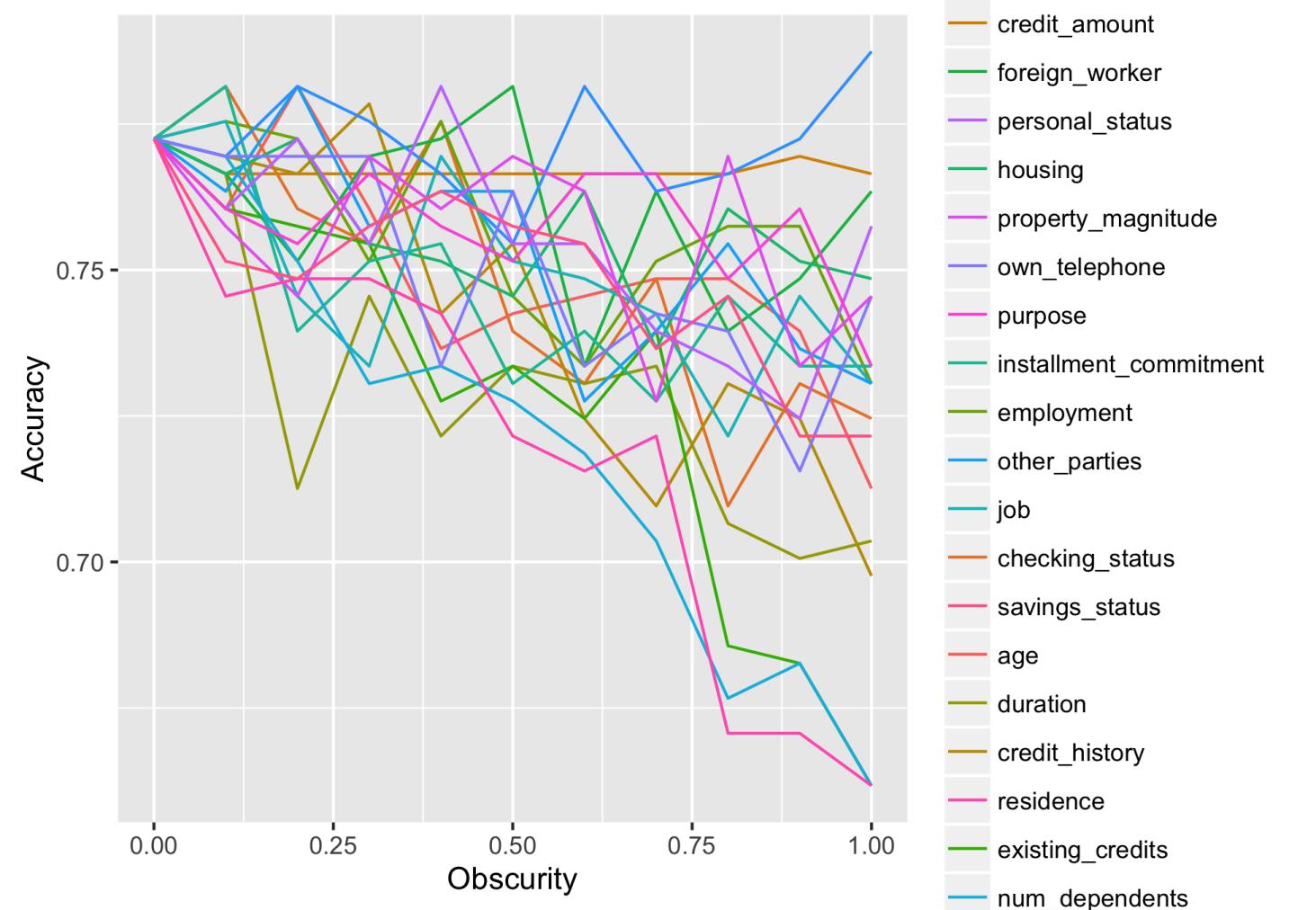} & \includegraphics[width=2in]{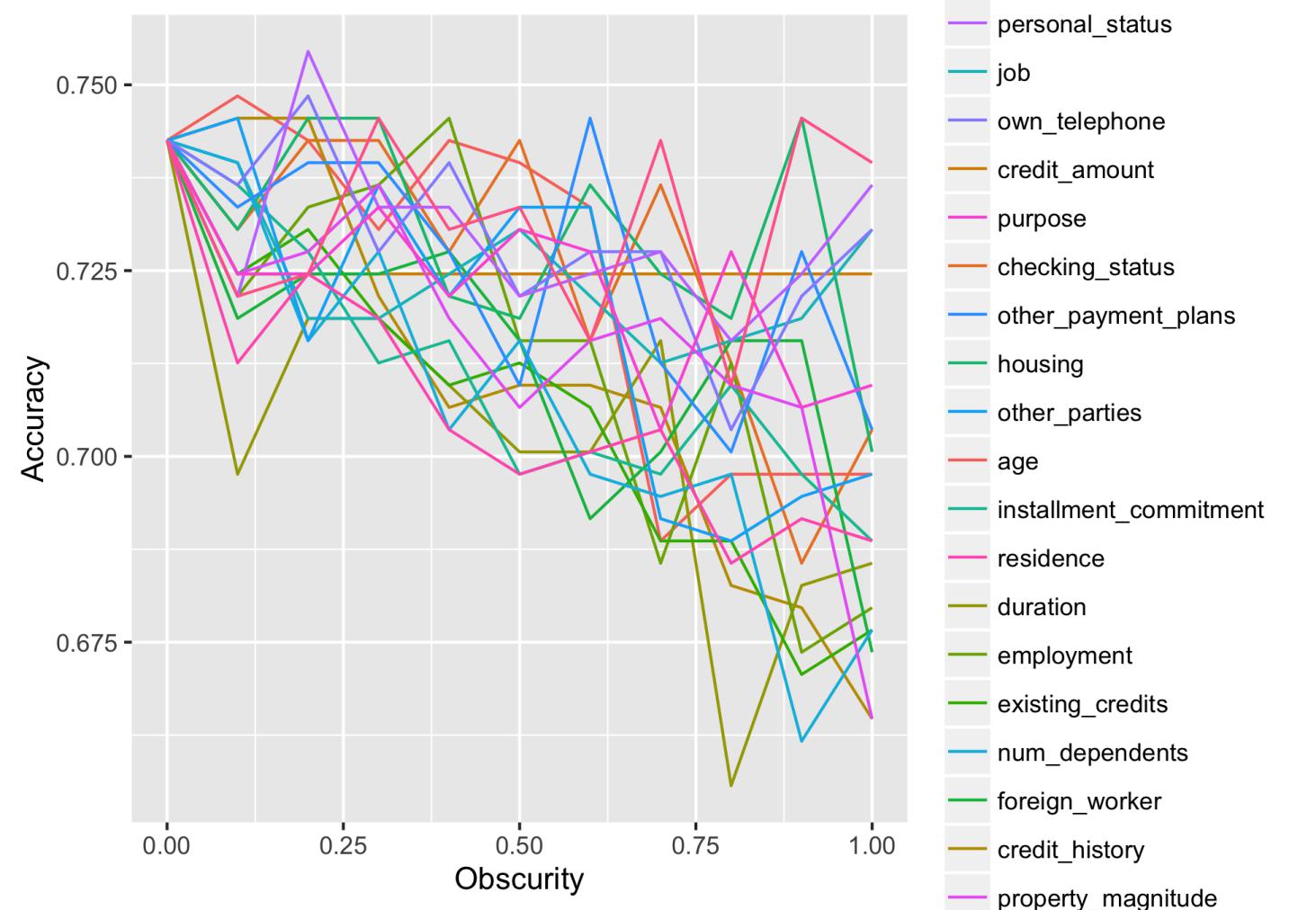}\\
\includegraphics[width=2in]{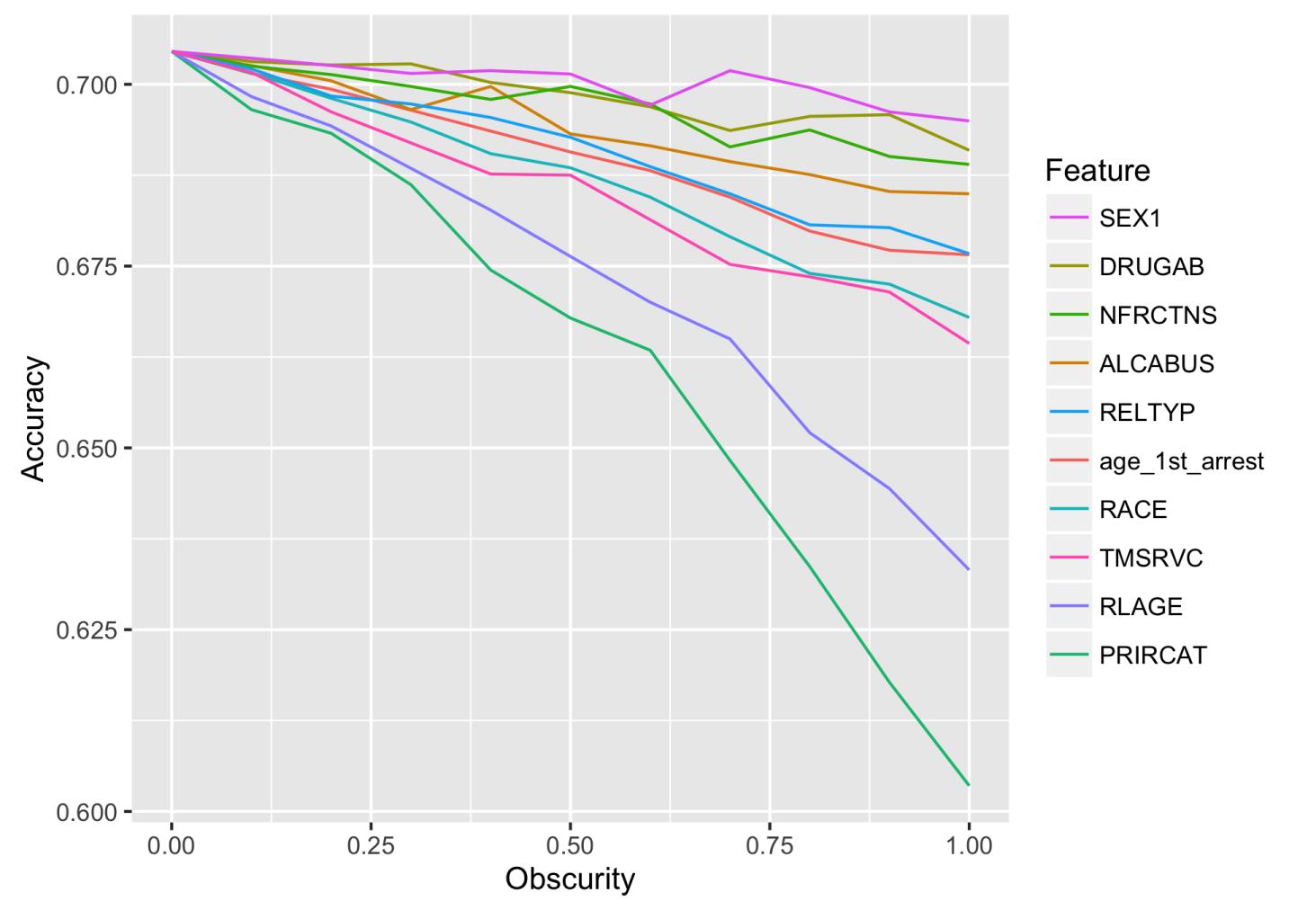} & \includegraphics[width=2in]{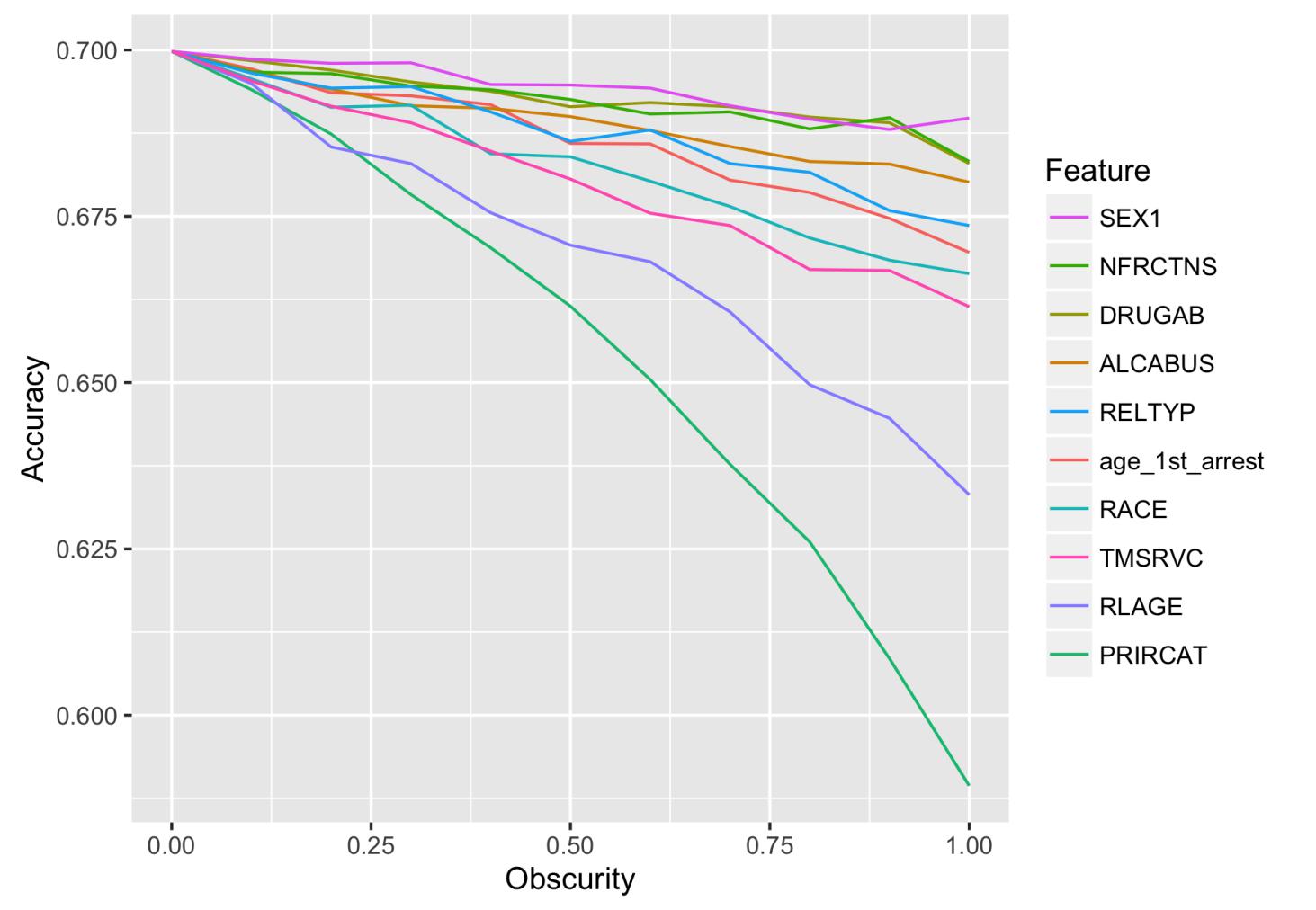} & \includegraphics[width=2in]{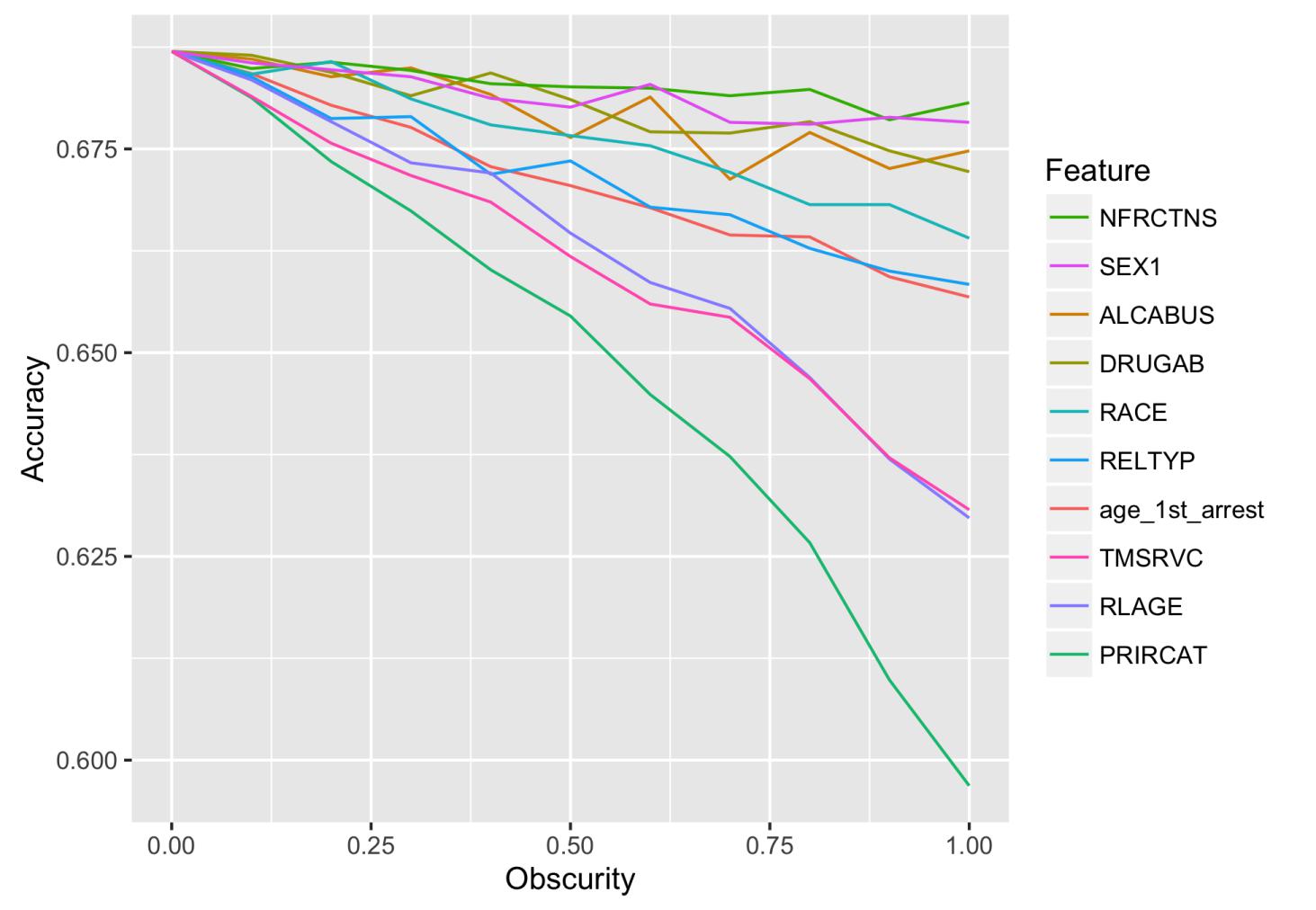}\\
\includegraphics[width=1.5in,height=1.5in]{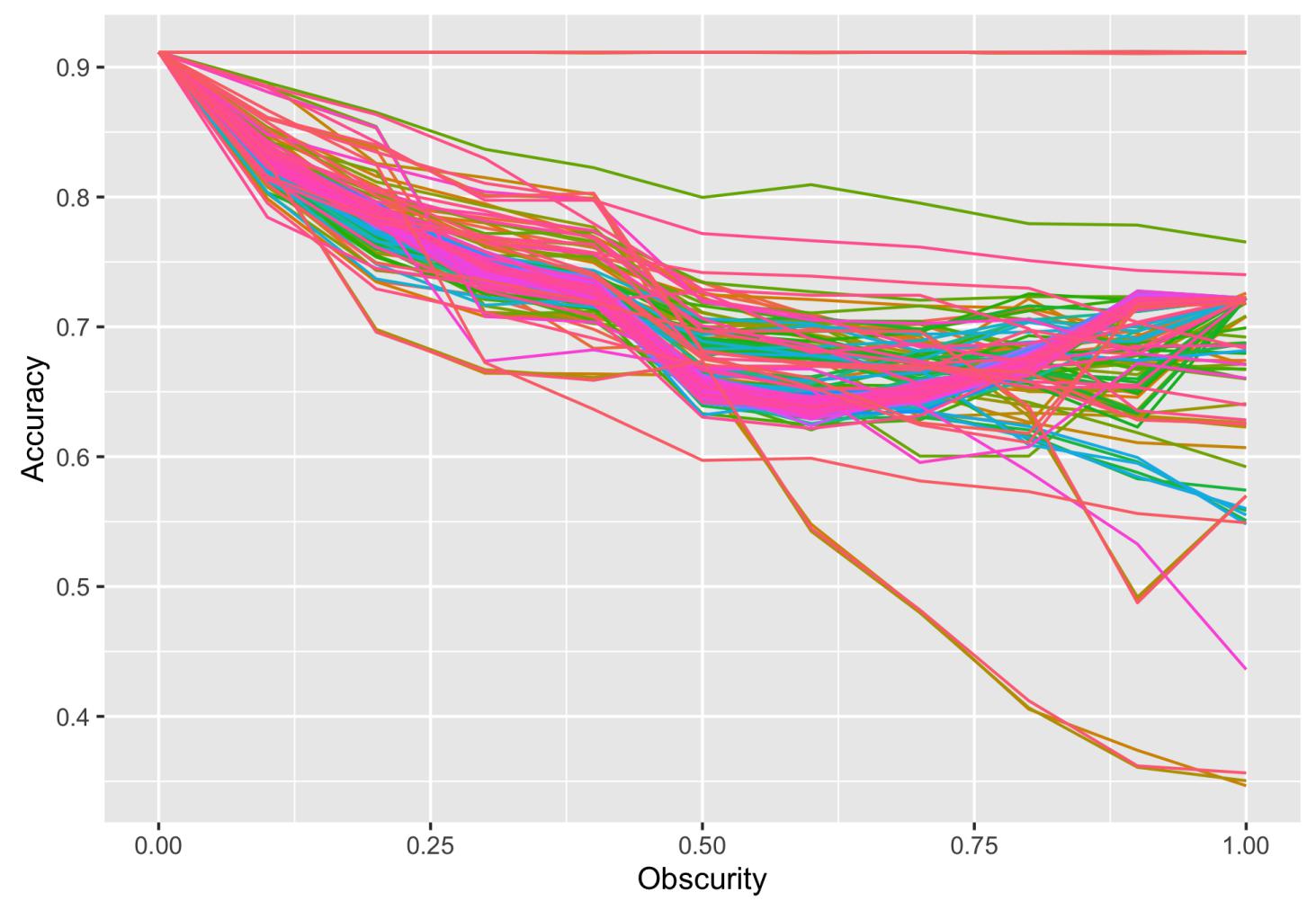}~~~~~~~~ & \includegraphics[width=1.5in,height=1.5in]{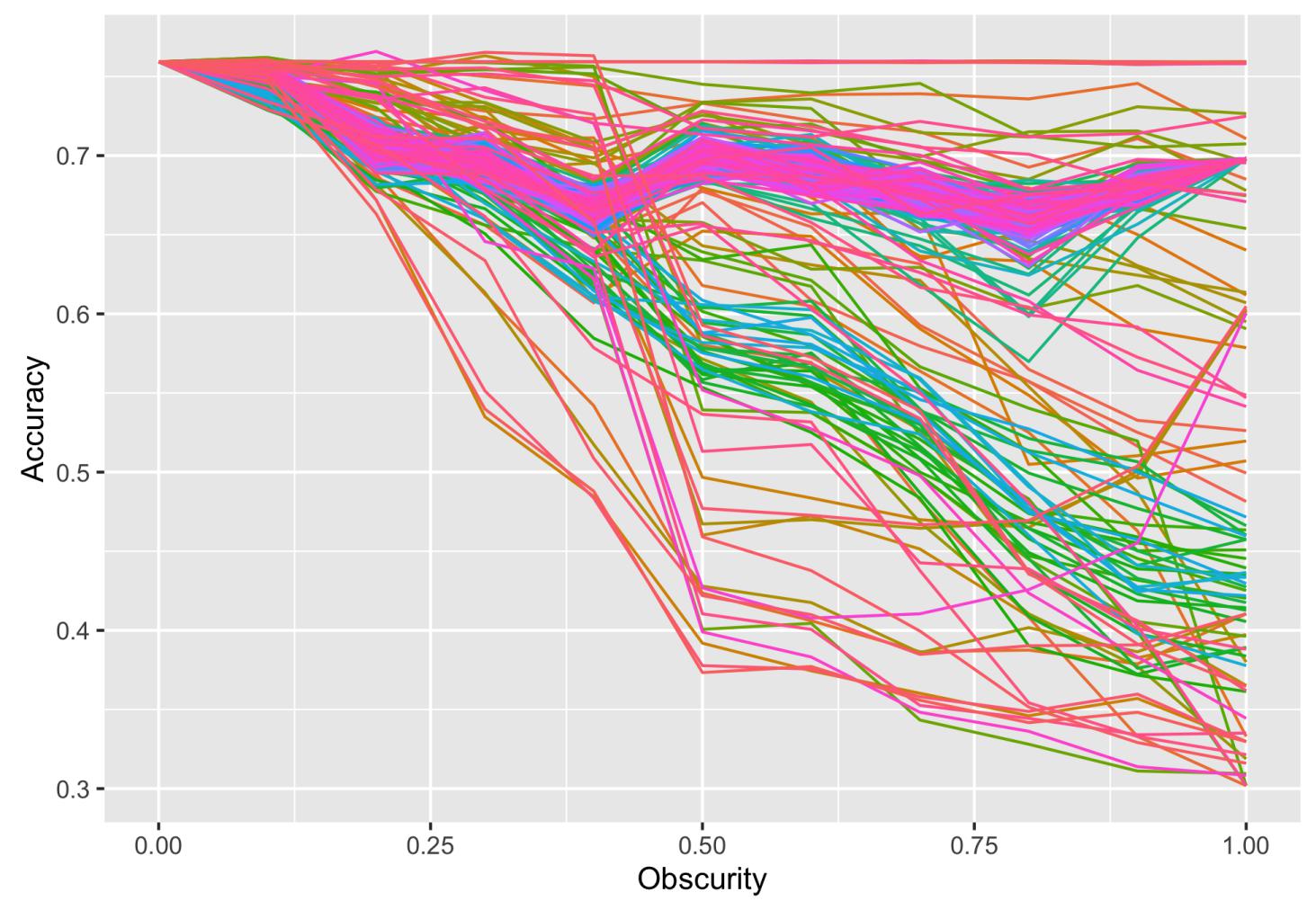}~~~~~~~~ & \includegraphics[width=1.5in,height=1.5in]{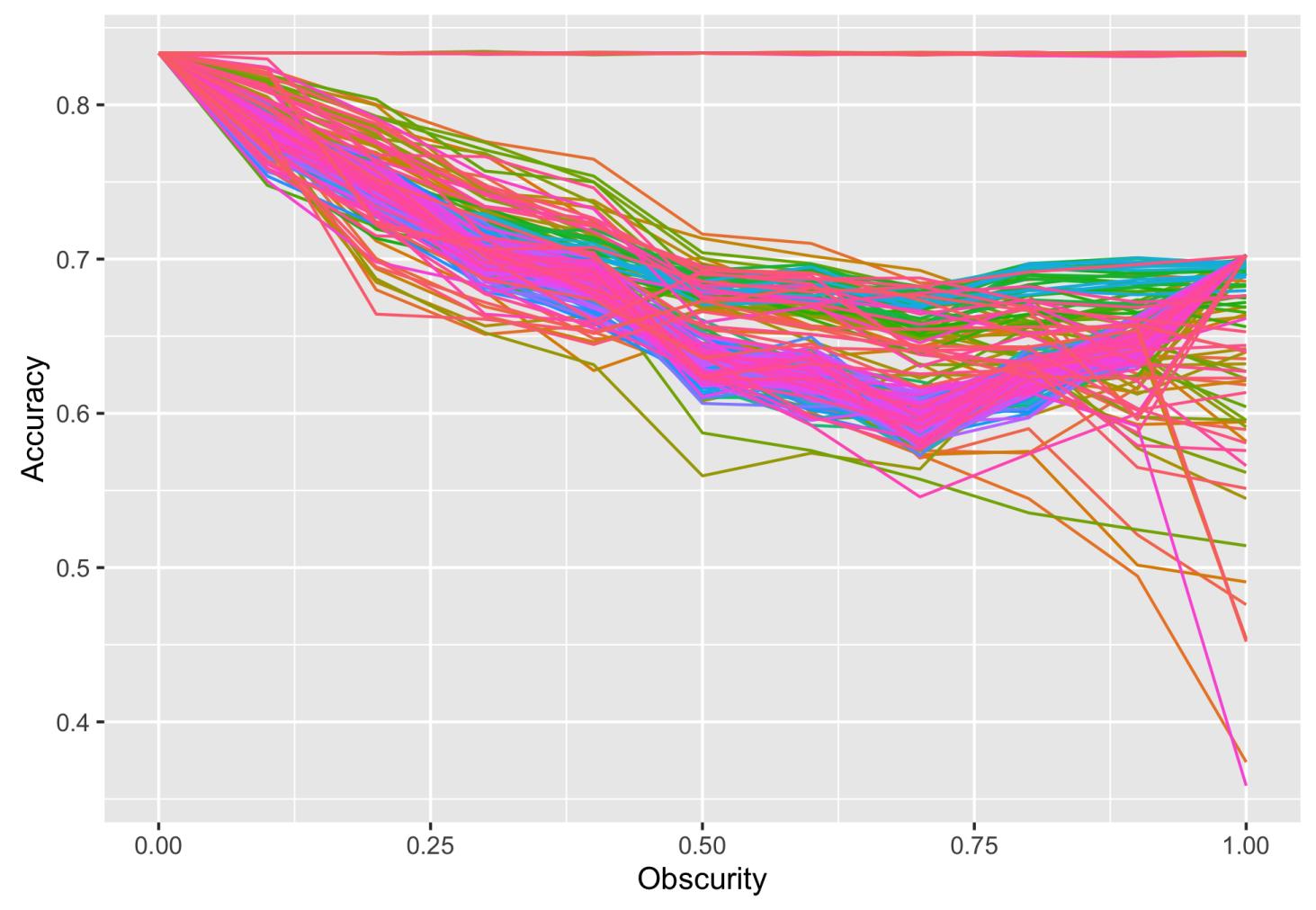}~~~~~~~~
\end{tabular}
\caption{Obscurity vs. accuracy plots for each model and each data set considered.  First column: C4.5 decision trees.  Second column: SVMs.  Third column: FNNs.  First row: Synthetic data.  Second row: Adult income data set.  Third row: German credit data.  Fourth row: Recidivism data.  Final row: Dark Reaction data, shown without a feature legend due to the large number of features.}
\label{fig:gfa_obscurity_acc}
\end{center}
\end{figure*}

We now assess the performance of our GFA method. 
We trained each model on each of the five data sets.  We then ran GFA using the test data for each data set. As we noted in Section~\ref{sec:gradient_auditing}, we  progressively increase the degree to which we obscure a data set by removing a variable. Specifically, we used partial obscuring values at $0.1$ intervals between $0$ (no removal) and $1$ (full removal) giving us 11 total partially obscured data sets to consider the accuracy change for. 
Figure \ref{fig:gfa_obscurity_acc} shows the resulting GFA plots.

\mypara{Synthetic data.} Beginning with the synthetic data under any of the models, we see that removing any one of the three main features (A, B, and C) that encode the outcome class causes the model to degrade to $50\%$ accuracy as our approach would predict.  Removing the constant feature has no effect on the model's accuracy, also as expected.  The removal of the random feature causing the model to lose a small amount of accuracy may initially seem surprising, however this is also as expected since the random feature also individually identifies each row and so could be used to accurately train an overfitted model.  

\mypara{Adult income data.}  On the Adult income data set, we see that the ranking changes depending on the model.  Recall that the more ``important,'' highly ranked features are those that cause the accuracy to drop the most, i.e. are towards the bottom of the charts.  While \texttt{race} is found to have only a small influence on all the models, the removal of \texttt{age} has a large impact on the SVM and FNN models, but is much less important to the decision tree.  On the FNN model gradient auditing plot we also see  a set of features which when partially removed actually \emph{increased} the accuracy of the model.  In a model that was not optimal to begin with, partially obscuring a feature may in effect reduce the noise of the feature and allow the model to perform better.

\mypara{German credit data.}  The results on the German credit data exhibit an arbitrary or noisy ordering.  We hypothesize that poor models are likely to produce such auditing results.  There are two interrelated reasons why this may be the case.  First, since the audit assesses the importance of a feature using the accuracy of the model, if the model is poor, the resolution of the audit is degraded.  Second, if the model is poor partially due to overfitting, obscuring features could cause spurious and arbitrary responses.  In these contexts, it makes more sense to consider the change in accuracy under a consistency measure.  We explore this further in Section \ref{sec:consistency}.

\mypara{Recidivism data.}  On the Recidivism data we see incredible consistency between the rankings of the different models.  The top three most important features under all three models are \texttt{PRIRCAT}, a categorical representation of the number of prior arrests of the prisoner, \texttt{RLAGE}, the age at release, and \texttt{TMSRVC}, the time served before the release in 1994.  The four least important features are also common to all three models: the sex of the prisoner, whether the prisoner is an alcohol or drug abuser, and the number of infractions the prisoner was disciplined for while incarcerated.

\begin{figure*}[!htb]
\begin{center}
\begin{tabular}{ccc}
\includegraphics[width=2in]{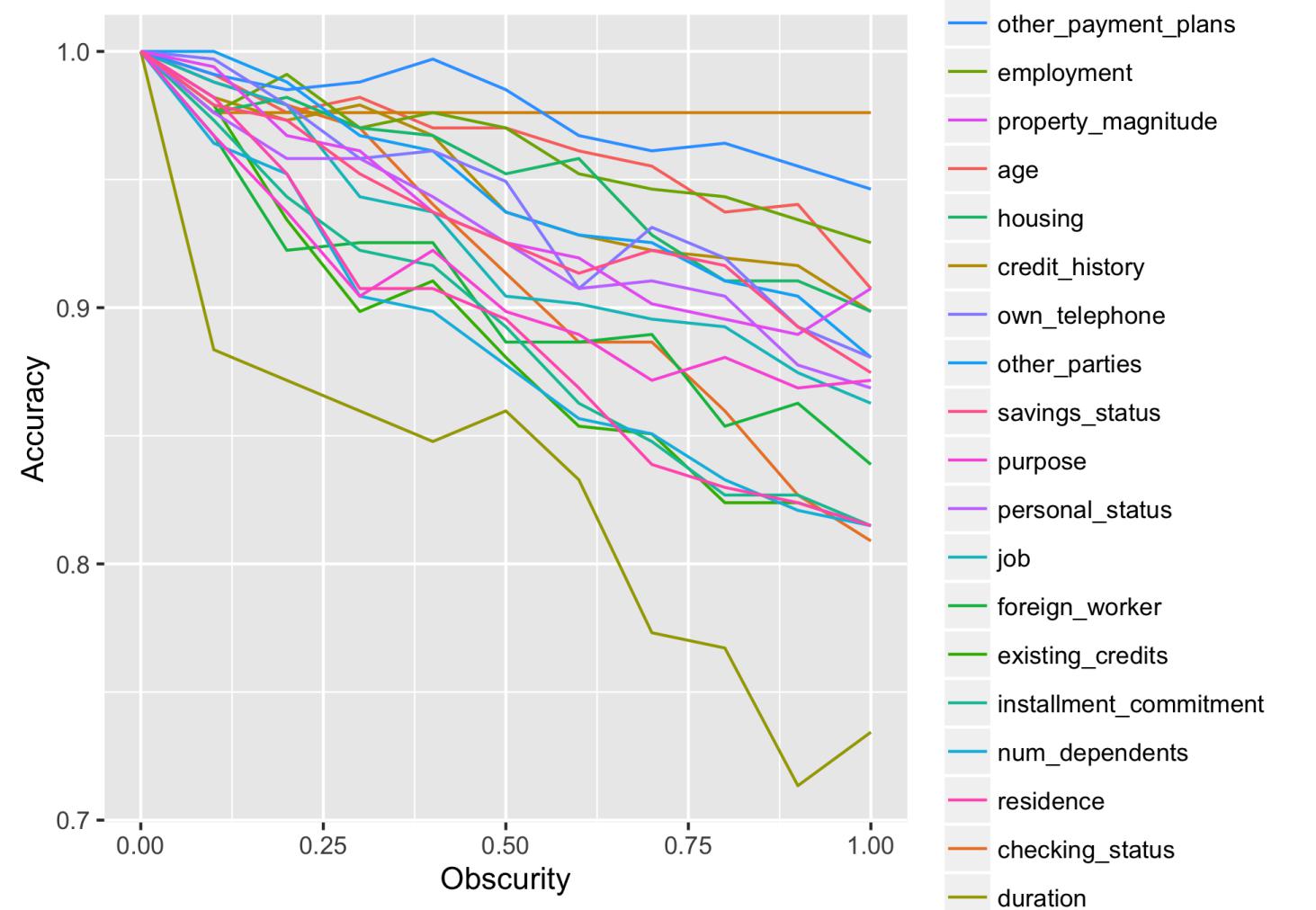} & \includegraphics[width=2in]{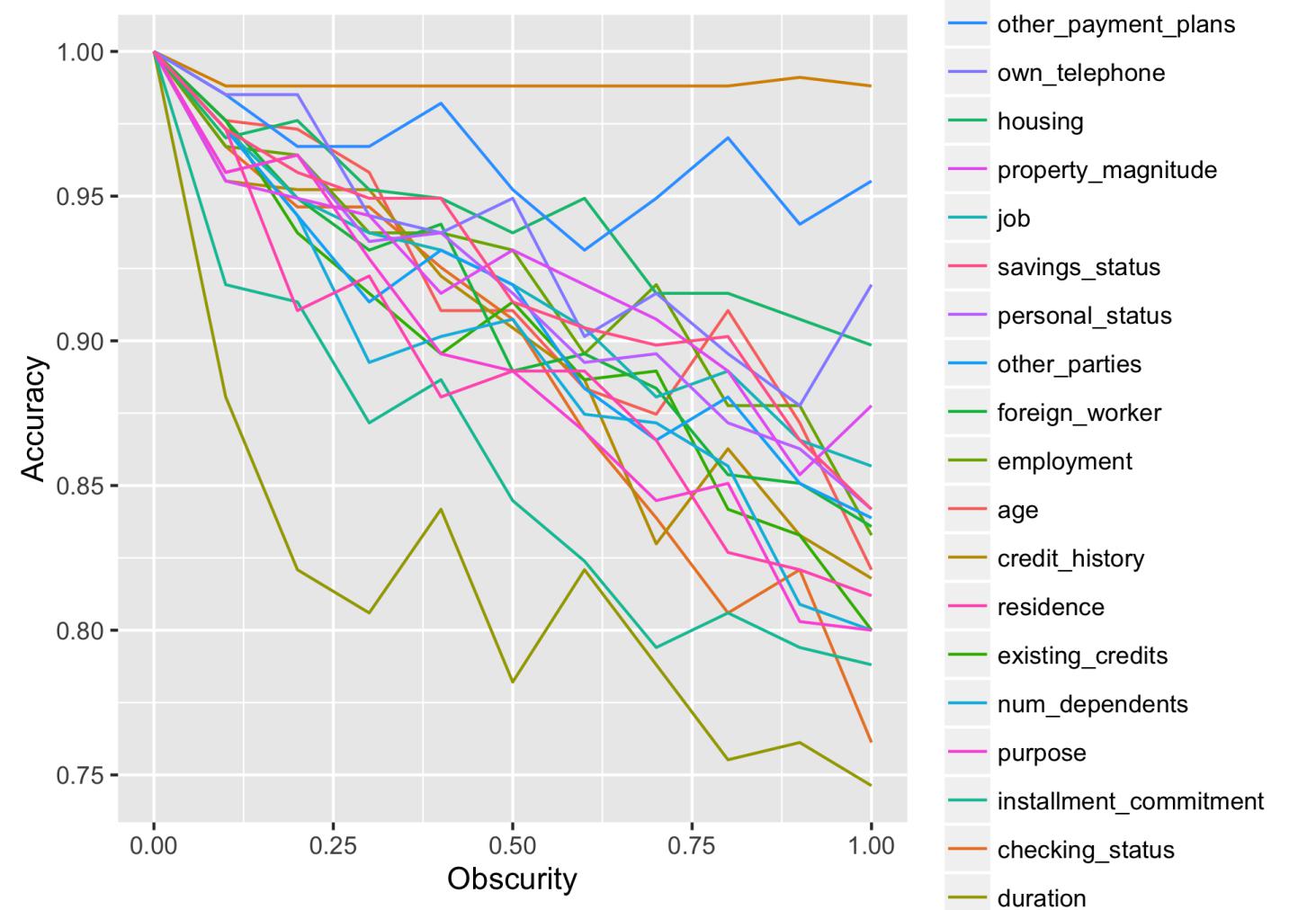} & \includegraphics[width=2in]{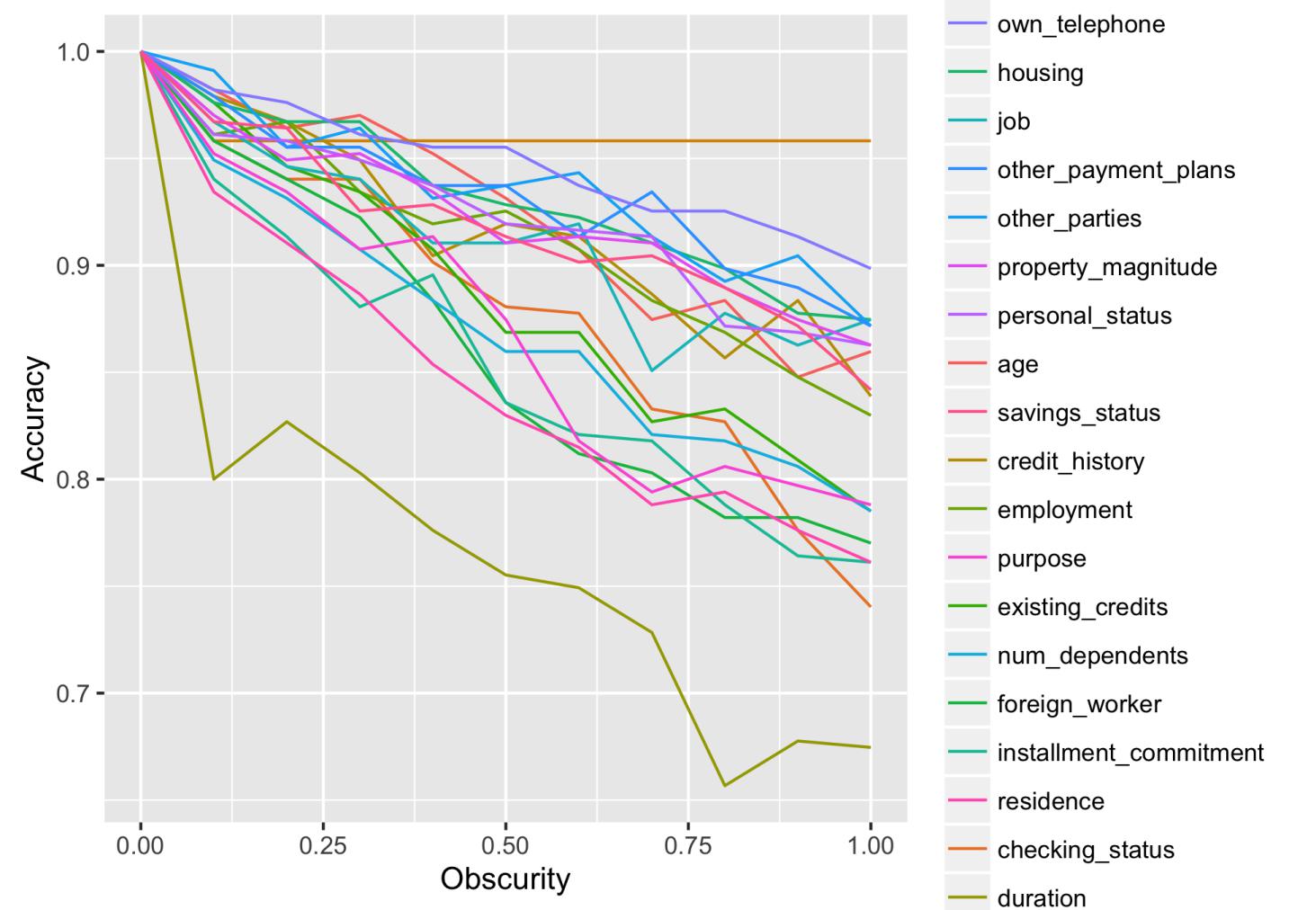}
\end{tabular}
\caption{Obscurity vs. consistency plots for the German Credit data.  First column: decision tree model. Second column: SVM.  Third column: FNN.}
\label{fig:german_consistency}
\end{center}
\end{figure*}

\mypara{Dark Reactions data.}  The Dark Reactions data shows different top ranked features for the three models, though all three rankings include the minimum Pauling electronegativity and the maximum Pearson electronegativity in the top ranked cluster of features. These values are calculated for the inorganic components of the reaction and have been shown to be important for distinguishing between chemical systems in this data set \cite{drpNature}, which this audit confirms.  Features indicating the presence of elements and amounts of metal elements with specific valence counts similarly allow the models to classify chemical systems. The SVM and FNN top features include atomic properties of the inorganics 
that are related to the electronegativity of an element,
so these proxies are correctly also highly ranked.  The top ranked decision tree features additionally include the average molecular polarizability for the organic components, which was previously hypothesized as important to synthesis of templated vanadium selenites explored via this data set \cite{drpNature}. 
For all three models, the lowest ranked descriptors are constants scored correctly as having no influence to the model.

\mypara{Running time.}
Running times for these experiments, including time to train the model, time to do all partially obscured audits, and time to write the partially obscured data sets to disk, varied from 13 seconds on the synthetic data set with the C4.5 decision tree model to just over 3 hours for the Dark Reaction data set with the FNN.  Since one-time audits are not highly time-sensitive tasks, and the code was unoptimized
, we present these times largely as evidence of the feasibility of this method.

\subsection{Auditing for consistency\label{sec:consistency}}

The results in Figure \ref{fig:gfa_obscurity_acc} allow us to both create a ranking of influence as well as evaluate absolute accuracy of the model on obscured data. However, 
the German Credit data set yields very  noisy results. To address this, we propose using \emph{model consistency}: we replace the original labels with those predicted by the model on unobscured data, and then calculate accuracy as we progressively obscure data with respect to these labels. The unobscured data will thus always have a consistency of 100\%.  The new gradient measured will be the difference between the 100\% consistency at obscurity of 0 and the degraded consistency when fully obscured.

As can be seen in Figure \ref{fig:german_consistency}, under the consistency measure the accuracy of the German Credit model degrades smoothly so that a ranking can be extracted.  The slight noise remaining is likely due to the final step of the categorical obscuring algorithm that redistributes ``lost observations" randomly.  The resulting ranking is fairly consistent across models, with \texttt{credit amount}, \texttt{checking status}, and \texttt{existing credits} ranked as the top three features in all models.

\begin{figure}[htbp]
\begin{center}
\includegraphics[width=3in]{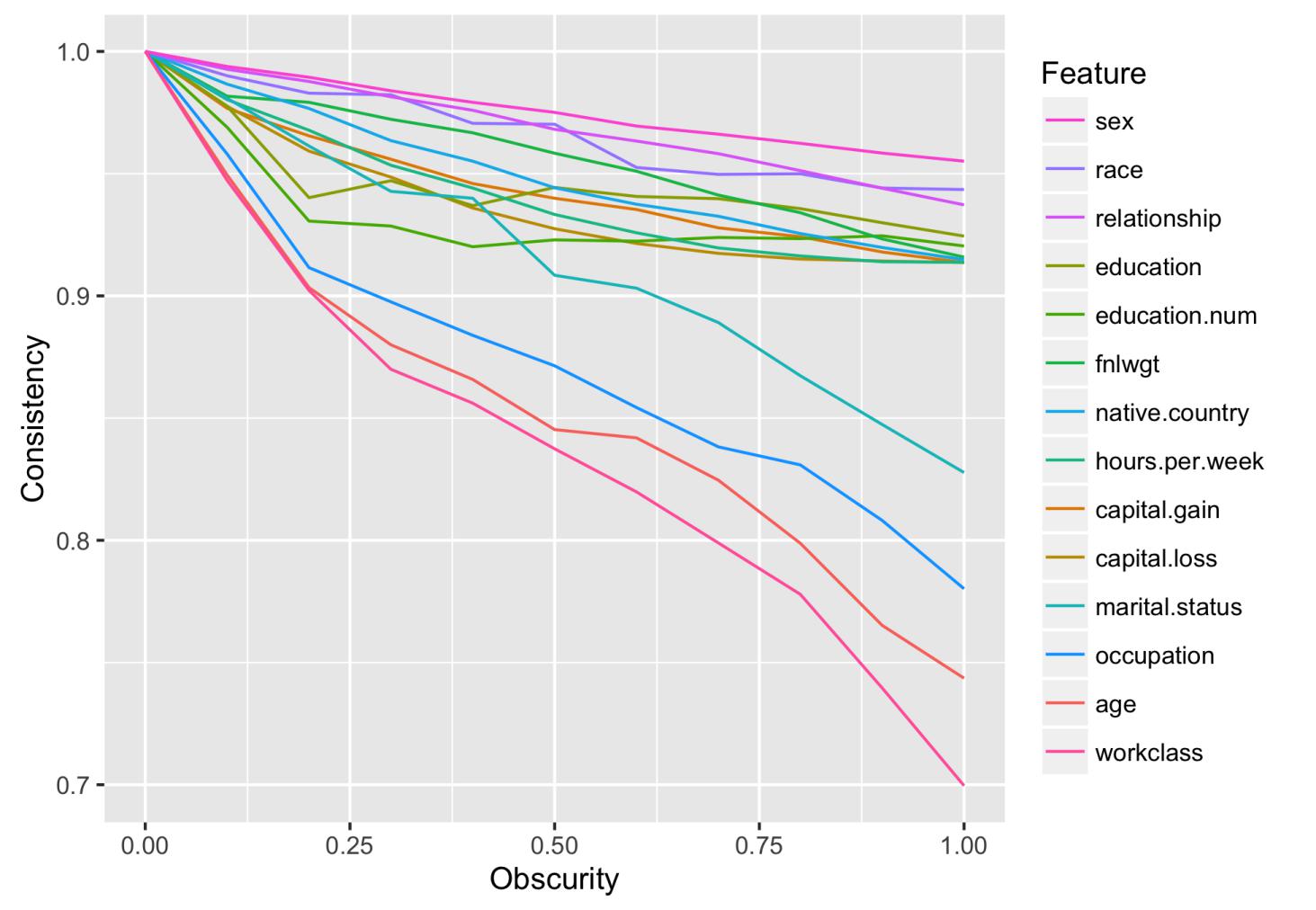}
\caption{Obscurity vs. consistency for Adult Income data modeled by an FNN.}
\label{fig:adult_consistency}
\end{center}
\end{figure}

Similar to the German Credit data, the FNN model on the Adult Income data set was not an optimal model.  This means that in the accuracy-based plots in Figure \ref{fig:gfa_obscurity_acc} obscuring the features at first leads, counterintuitively, to an increase in accuracy.  In the consistency graph for the FNN model on the Adult Income data (see Figure \ref{fig:adult_consistency}) we see that while the ranking derived from accuracy closely resembles the consistency ranking, a cluster of features (\texttt{native.country}, \texttt{capital.loss}, \texttt{capital.gain}, and \texttt{hours.per.week}) had been ranked above \texttt{occupation} and \texttt{marital.status} and the consistency ranking moves that cluster down in rank.

\subsection{Evaluating with respect to a direct influence audit}

In order to determine if GFA is correctly determining the indirect influence of a model, we will first develop a simple method of detecting \emph{direct} influence, and then examine the outlying attributes for which our indirect audit differs.

\mypara{Direct influence audit.}  The first step of the direct influence audit method we will use is the creation of an interpretable model of the model.  By a ``model of a model'' we mean that we should 1) train the model $f$ on training data $(X, Y)$, 2) determine new labels $\hat{Y}$ from the predicted outcomes $f(X)$, and 3) overfit an interpretable model $I(f)$ to these predicted labels (as done in \cite{Barakat2004Learning}).  (This idea is similar to model compression, but without the need to find new test data \cite{bucilua2006model}.)  Assuming that the model resulting from this procedure has high accuracy on $\hat{Y}$, we now have an interpretable model of our model.

For the SVM and decision tree models trained on the each of the five data sets, we trained an unpruned C4.5 decision tree model of the model.  With these interpretable models of a model, unfortunately a manual comparison to the feature ranking is still impossible due to the size of the resulting trees.  We create feature importance scores by calculating the probability that each feature appeared on the path from the root to the leaf node containing an item from the training set.  This ranking is weighted by the number of items at each leaf node.  Any feature appearing at the root node, for example, will have a probability of 1 and come first in the derived ranking.  This gives us the direct influence audit results we will compare to.

\mypara{Synthetic data.}  Beginning with the simple decision tree model for the synthetic data, looking at the decision tree reveals that only feature A is used explicitly - the decision tree has a single split node.  When we create a decision tree model of this model, to confirm the model of a model technique, the result is exactly the same decision tree.  Creating a decision tree model of the SVM model, we find again that there is a single node splitting on feature A.  Both models of models have 100\% accuracy on the training set (that they were purposefully over-fit to).  The probability ranking of all of these models for the synthetic data contains feature A first with a probability of 1 and all remaining features tied after it with a probability of 0 (since only feature A appears in the decision tree).  Since the probability ranking is a direct audit, it is not surprising that it does not score proxy variables B and C highly.

\mypara{Comparison to a direct influence audit.} 
To evaluate the model of a model probability ranking in comparison to the GFA accuracy ranking, we compare the feature rankings on three datasets (Adult, Recidivism and German).  Specifically, we collect the three generated rankings into one vector for each of the feature ranking procedures and run a Spearman rank-correlation statistical test. We find a combined sample rank correlation of $0.413$ (a 95\% bootstrap confidence interval of $[0.115, 0.658]$), and find also that we can reject the null hypothesis (of no correlation), with $p<0.002$.  When we compare the GFA ranking based on model \emph{consistency} (cf. Section~\ref{sec:consistency}), the results are similar to the ones based on model accuracy.  This provides evidence that our feature auditing procedure closely matches a direct influence audit overall.

\begin{figure}[htbp]
\begin{center}
\includegraphics[width=3.5in]{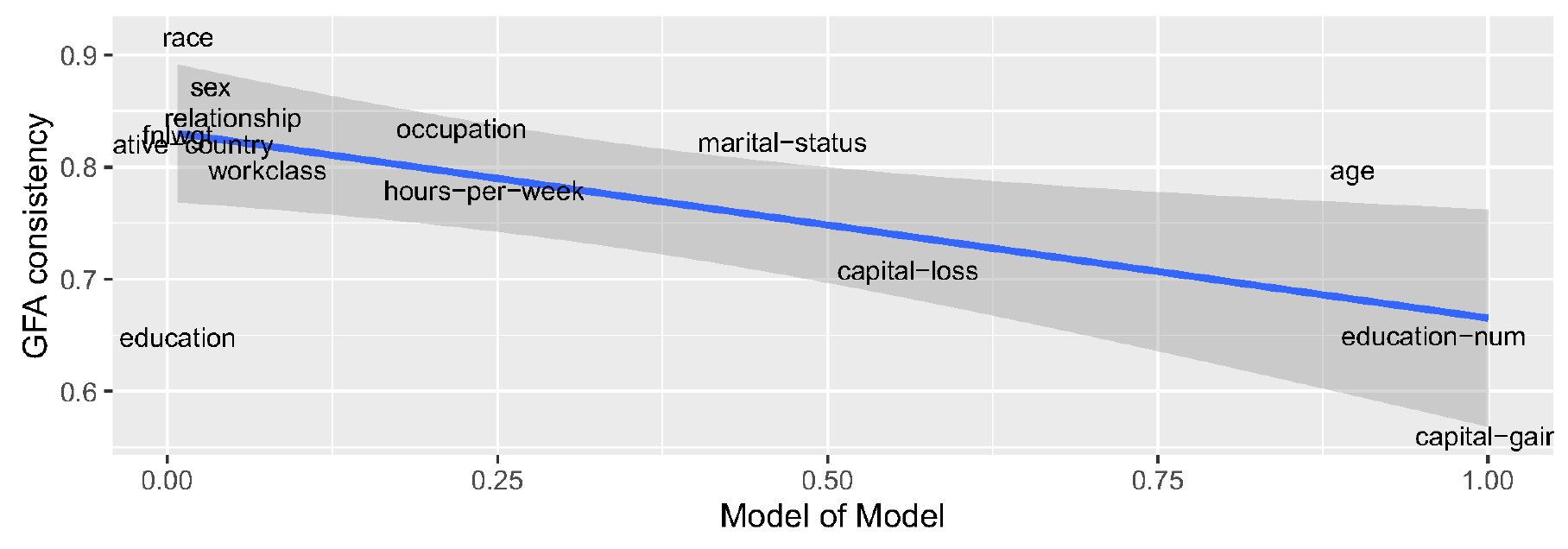}
\caption{Model of a model decision tree probability rankings vs. GFA consistency scores shown with a linear regression and 95\% confidence interval.  The outlying features contain proxy information in the data set.}
\label{fig:linreg_mom}
\end{center}
\end{figure}

To consider the cases where these rankings don't match, we look at Adult Income under the C4.5 decision tree model.  As shown in Figure \ref{fig:linreg_mom}, linear regression confirms that most features have similar scores, but there are a few important outliers: \texttt{marital-status}, \texttt{education}, \texttt{race}, \texttt{age}, and \texttt{capital-gain}.  We hypothesize that the information in these features can be reconstructed from the remaining attributes, and so they are scored differently under an indirect influence audit.  We explore this hypothesis next.

\subsection{Comparison to previous work}
\label{sec:comparison}

Henelius et al. \cite{Henelius2014BlackBox} provide a different technique to solve the Black-box Feature Auditing problem.  They focus on determining not only the influence scores associated with each feature, but also groupings of features that are more influential as a group than they are individually (i.e., have mutual influence on a model's outcomes) and use the consistency measure as the score.  The results of running their algorithm\footnote{Available at: https://bitbucket.org/aheneliu/goldeneye/} on the synthetic data we considered here is shown in Table \ref{table:comparison}.

\begin{table}[ht]
\begin{center}
\begin{tabular}{l|ll|ll}
&\multicolumn{2}{c}{C4.5 Decision Tree} & \multicolumn{2}{c}{SVM} \\
Feature & Henelius et al.  & GFA & Henelius et al.  & GFA \\
\hline
A & 0.50 & 0.50 & 0.87 & 0.50 \\
B &    1.0 & 0.53 & 0.87 & 0.50  \\
C & 1.0 & 0.53  & 0.88 &  0.50 \\ 
Random &  1.0 & 0.96 &   0.99 & 0.97 \\
Constant &  1.0 & 1.0 &1.0 & 1.0 \\
\hline

\end{tabular}
\end{center}
\caption{\normalfont Synthetic data comparison between Henelius et al. and GFA.  All models achieve 100\% accuracy, so GFA consistency and accuracy are the same.}
\label{table:comparison}
\end{table}%

These scores on the synthetic data set illuminate the key difference between the Henelius et al. work and GFA: while Henelius et al. focus on auditing for \emph{direct influence} on a model, GFA includes both direct and \emph{indirect influence} through proxy variables.  For example, the C4.5 decision tree created on the synthetic data is a very simple tree containing one node that splits on the value of feature A.  Henelius et al. correctly show that A is the only feature directly used by the model.  However, GFA additionally shows that features B and C are proxies for A (recall that B is defined as two times feature A and C is defined as negative one times feature A).

\begin{table}[ht]
\begin{center}
\begin{tabular}{l|lll|ll}
&\multicolumn{3}{c}{C4.5 Audit Scores} & \multicolumn{2}{c}{Feature predictability} \\
Feature & Henelius & GFA  & GFA  & REPTree & C4.5 or \\
& et al. & cons. & acc. & & M5\\
\hline
capital-gain & 0.94 & 0.56 & 0.55 &0.16 & 0.21 \\
education & 0.98 & 0.65 & 0.65 & 1.0 & 1.0\\
education-num & 0.92 & 0.65 & 0.65 & 1.0 & 1.0 \\
capital-loss & 0.98 & 0.71 & 0.68 & 0.09 & 0.16\\
hrs-per-week & 0.96 & 0.78 & 0.74 & 0.44 & 0.49\\
age & 0.94 & 0.80 & 0.76 & 0.65 & 0.68\\
workclass & 0.98 & 0.80 & 0.76 & 0.11 & 0.16\\
fnlwgt & 0.99 & 0.83 & 0.77 & 0.15 & 0.22\\
marital-status & 0.87 & 0.82 & 0.77 & 0.74 & 0.76\\
native-country & 1.0 & 0.82 & 0.78 & 0.22 & 0.28\\
occupation & 0.90 & 0.84 & 0.79 & 0.21 & 0.17 \\
relationship & 0.99 & 0.84 & 0.81 & 0.69 & 0.70 \\
sex & 1.0 & 0.87 & 0.82 &0.62 & 0.62 \\
race & 1.0 & 0.92 & 0.83 & 0.25 & 0.23\\
\hline

\end{tabular}
\end{center}
\caption{\normalfont Adult Income data comparison between Henelius et al and GFA consistency and accuracy scores for a C4.5 decision tree model.  Feature predictability scores are correlation coefficient or Kappa statistic (for numerical or categorical features, respectively) when predicting that feature from the remaining features using two tree-based models.
}
\label{table:comparison_adult}
\end{table}%

\begin{figure}[htbp]
\begin{center}
\includegraphics[width=3.5in]{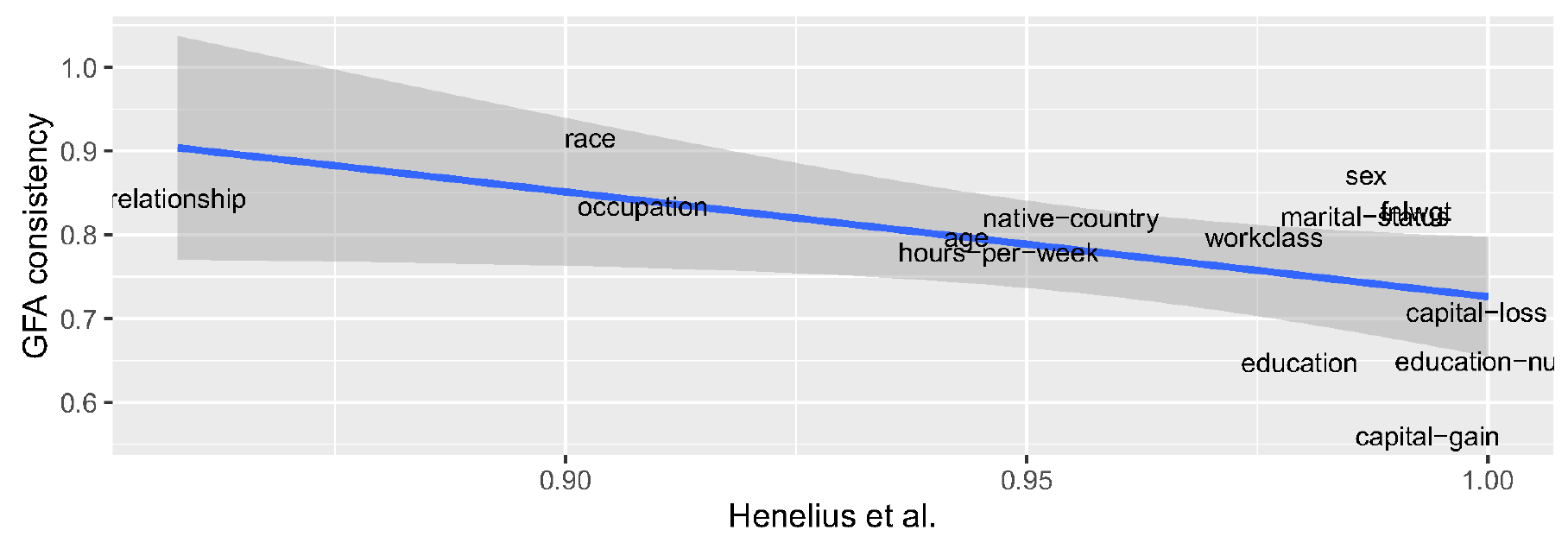}
\caption{Henelius et al. influence scores vs. GFA consistency scores shown with a linear regression and 95\% confidence interval.  The outlying features contain proxy information in the data set.}
\label{fig:linreg_henelius}
\end{center}
\end{figure}

For a real-world comparison to Henelius et al., we consider the Adult data set under a C4.5 decision tree model.  The scores and rankings generated by Henelius et al. do not match those generated by GFA.  Figure \ref{fig:linreg_henelius} shows that features \texttt{marital-status}, \texttt{fnlwgt}, \texttt{sex}, \texttt{education}, \texttt{education-num}, and \texttt{capital-gain} are outliers.  In order to determine if this is due to the presence of proxy variables, or variables that are more complexly encoded in the remaining attributes by the decision tree, we then used two tree-based models to predict each feature from the remaining features (see Table \ref{table:comparison_adult}).\footnote{Weka's REPTree, J48, and M5P models were used for this analysis with the default model-building parameters.  J48 was used to predict categorical features and M5P was used for numerical features.  REPTree can handle both categorical and numerical features.}  Models were built on the test set (using a $\frac{2}{3}:\frac{1}{3}$ training-test split) in order to replicate the data used for the audit.  Reported predictability scores are the correlation coefficient for numerical features and the Kappa statistic for categorical features.

Looking at the resulting feature predictability scores, we see that \texttt{education} and \texttt{education-num} are both perfectly reconstructable from the remaining attributes.  This is not surprising since \texttt{education-num} is a numerical representation of \texttt{education} and thus, a perfect proxy.  The GFA consistency and accuracy measures both have \texttt{education} and \texttt{education-num} as tied for an importance score of $0.65$, thus confirming that GFA handles indirect influence, while Henelius et al. has \texttt{education} with a score of $0.98$ while \texttt{education-num} scores $0.92$.

The features \texttt{marital-status} and \texttt{relationship} are both highly, but not exactly, predictable from the remaining attributes.  This is likely because these are close, but not exact, proxies for each other.  For example, the relationship status ``Unmarried" may mean a marital status of ``Divorced," ``Never-married," or ``Widowed."  The GFA scores for \texttt{marital-status} and \texttt{relationship} show they are of similar importance to the model, with consistency scores of $0.82$ and $0.84$ respectively (ranked 8.5 and 11.5 in importance), and accuracy scores of $0.77$ and $0.81$ (ranked 8.5 and 12).  The Henelius et al. scores are less similar at $0.87$ (ranked most important) and $0.99$ (ranked 10.5).  The GFA closely matching scores shows the procedure accounts for indirect influence of a feature, while Henelius et al. does not.

Recent work by Datta et al. \cite{datta2016} (discussed in more depth in Section~\ref{sec:altern-appr}) present a similar approach, focusing on direct influence, that can additionally identify proxy variables.  Identifying the influence of proxy variables using this method proceeds in two steps: first, the proxy variables are identified, then their individual direct influence in the ranked list is found.  Considering the Adult data set, but under a different model, they find that \texttt{marital-status} and \texttt{relationship} are both proxies for \texttt{sex}.  Their ranking finds that \texttt{marital-status} is ranked first for influence, \texttt{relationship} third, and \texttt{sex} ninth.  Additionally, their influence score for \texttt{relationship} is under half of that for \texttt{marital-status}, and the influence score for \texttt{sex} is under half of that for \texttt{relationship}.  Thus, similar to Henelius et al., the Datta et al. two step procedure does not account for the shared indirect influence of \texttt{marital-status} and \texttt{relationship} on the outcome.

\subsection{Comparison to feature selection\label{sec:featureselection}}
Finally, we examine the relation between our auditing procedure and feature selection methods. While GFA is fundamentally different from feature selection since\begin{inparaenum}[a)]
\item features may not be removed from the model, and 
\item the model may not be retrained,
\end{inparaenum}
due to their similar ranked feature results, we compare the resulting GFA ranking to a feature selection generated ranking.  Feature selection was applied to the synthetic, Adult Income, and German Credit data sets.  It was performed using a wrapper method (around both C4.5 decision trees and SVMs) and a greedy forward search through attribute subsets to generate a ranking.\footnote{Feature selection was implemented in Weka version 3.6.13 using WrapperSubsetEval and Greedy StepWise on J48 and SMO models.  Default options were used, save for the generation of a complete ranking for all features.}
For both the Adult Income and German Credit data sets feature selection created identical rankings for both decision trees and SVMs.

Spearman's rank correlation is a nonparametric test of the relationship between two rankings, in this case the rank orderings generated by feature selection and by GFA.  The synthetic data had a strong correlation between these rankings, with feature A ranked first by all methods, though given the small feature size this is not statistically significant.  The Adult Income and German Credit data set ranking comparison correlations were both weak.  This was not improved when using the consistency feature ranking instead, and in the case of the C4.5 decision tree was an order of magnitude weaker.

The weak correlation is not surprising since the two methods ask philosophically different questions.  We are interested in the importance of a feature to a specific instance of a model, while feature selection considers importance with respect to an as-yet uninstantiated model. In addition, feature selection gives an indication of the direct influence of a feature, while GFA is focused on indirect influence.


\section{Discussion\label{sec:discussion}}

Feature influence is a function of the interaction between model and data. A feature may be informative but not used by the classifier, or conversely might be a proxy and still useful. Thus, influence computation must exploit the interaction between model and data carefully.  
If the obscured and unobscured datasets are similar, then the classifier can't have found useful signal and the classifier's outputs won't change much under our audit.
If there \emph{were} significant differences and the classifier used these differences in its model, then gradient feature auditing will show a change in the classifier behavior, as desired. Finally, if there were differences between attribute subgroups but those differences are irrelevant for the classifier, then gradient feature auditing will not show a large change in classifier behavior. This ``sensitivity to irrelevance'' is an important feature of a good auditing procedure.

It remains a challenge to effectively compare different approaches for auditing models since, as we have seen, different approaches can have points of agreement and disagreement. Our obscuring procedure prefers to use a computational metaphor -- predictive power -- rather than a statistical metaphor such as hypothesis testing, but it seems likely that there are ways to relate these notions. Doing so would provide a combined mathematical and computational framework for evaluating black-box models and might help unify the different existing approaches to performing audits.

\section{Related Work}
\label{sec:related-work}
 
In addition to early work by Breiman \cite{breiman2001random} and the recent works by Henelius et al. \cite{Henelius2014BlackBox} and Datta et al. \cite{datta2016}, a number of other works have looked at black-box auditing, primarily for direct influence  \cite{Duivesteijn2014,zacarias2013comparing,strobl2007bias,strobl2008conditional}.
There are potential connections to privacy-preserving data mining \cite{agrawal2000privacy}: however, the trust model is different: in privacy-preserving data mining, we do not trust the user of the results of classification with sensitive information from the input. In our setting, the ``sensitive'' information must be hidden from the classifier itself. Another related framework is that of \emph{leakage} in data mining \cite{kaufman2012leakage}, which investigates whether the methodology of the mining process is allowing information to leak from test data into the model. One might imagine our obscuring process as a way to prevent this: however, it might impair the efficacy of the classifier.

Another related topic is \emph{feature selection}, as we discussed in Section~\ref{sec:featureselection}. 
From a technical standpoint (see  \cite{Chandrashekar2014FeatureSelection}), the \emph{wrapper} approach to feature selection (in which features are evaluated based on the quality of the resulting prediction) is most related to our work. One such method is stepwise linear regression, in which features are removed from input for a generalized linear model based on their degree of influence on the model to make accurate predictions, measured using a correlation coefficient.  

Model \emph{interpretability} focuses on creating models that are sparse, meaningful, and intuitive and thus human-understandable by experts in the field the data is derived from \cite{Ustun2014SLIM}.  The classic example of such models are decision trees \cite{Quinlan1993C4.5},
while recently supersparse linear integer models (SLIMs) have been developed as an alternative interpretable model for classification \cite{Ustun2014SLIM}.  New work by Ribeiro et al. \cite{lime} trains a shadow interpretable model to match the results of a given classifier.  For neural networks, various approaches based on visualizing the behavior of neurons and inputs have also been studied \cite{Zeiler2014Visualizing,kabra2015understanding,Le2013Building}.

Finally, we note that a compelling application for black-box audits of indirect influence includes considerations of algorithmic fairness \cite{Romei13Multidisciplinary}. Understanding the influence of variables can help with such a determination.


\raggedright
\bibliographystyle{abbrv}
\bibliography{audits}
\end{document}